%% file: main.tex
\documentclass[runningheads]{llncs}

\usepackage{eccv}

\usepackage{eccvabbrv}

\usepackage{graphicx}
\usepackage{booktabs}
\usepackage{amsmath}
\usepackage{amssymb}
\usepackage{algorithm}
\usepackage{algorithmic}
\usepackage{float}
\usepackage{multirow}
\usepackage{microtype}
\newcommand{\Romannum}[1]{\MakeUppercase{\romannumeral #1}}

\usepackage[accsupp]{axessibility}  

\usepackage{hyperref}

\usepackage{orcidlink}

\begin{document}
\emergencystretch=1em

\title{\texorpdfstring{JECA$^2$: Judgment--Explanation Consistent Adversarial Attack against\\ Forensic Vision-Language Models}{JECA2: Judgment--Explanation Consistent Adversarial Attack against Forensic Vision-Language Models}}

\titlerunning{JECA$^2$: Judgment--Explanation Consistent Adversarial Attack}
\author{Jiachen Qian}

\authorrunning{Qian}

\institute{City University of Hong Kong
\email{72510756@cityu-dg.edu.cn}}

\maketitle

\begin{abstract}
Image forensics aims to determine the authenticity of visual content. Recent methods use vision–language models (VLMs) to conduct forensic analysis, achieving excellent detection accuracy and providing natural-language explanations for their judgments to support downstream human-assisted review.
Despite these advantages, it remains unclear whether forensic VLMs can resist consistency-oriented adversarial stress tests, which is important for deployment auditing. To identify potential flaws, we propose a Judgment-Explanation Consistent Adversarial Attack (JECA$^2$) tailored for forensic VLMs. Unlike attacks that mainly aim to mislead a target model into making an incorrect prediction, JECA$^2$ induces a forensic VLM to generate both an incorrect judgment and an explanation that supports its faulty judgment under an automated consistency metric, which may make downstream explanation-based triage more difficult.
Experiments under controlled white-box red-team protocols show that JECA$^2$ achieves a high attack success rate and high automated judgment-explanation consistency in the primary diagnostic setting. We further evaluate image-only, predicted-mask, and transfer settings to characterize how the attack changes under weaker threat conditions.


\end{abstract}

\section{Introduction}
\label{sec:intro}



Riding the wave of rapid progress in generative models, image editing technologies such as image inpainting~\cite{b80} and attribute editing~\cite{b51}, have flourished over the past years. They enable users to generate realistic, manipulated images, which boost industries such as artistic creation and product design, yet posing severe threats to digital trust. A countermeasure to mitigate the risk is image forensic, which typically train discriminative models to classify an image as real or fake~\cite{b85,b60}, or to localize manipulated regions via pixel-level supervision~\cite{b1}.

Like many fields in signal and image processing, image forensic is being reshaped by the development of large models. Cutting-edge methods~\cite{b6,b7,bForgeryGPT,bFFAA} supervisedly fine-tune foundation vision-language models (VLMs) on massive collections of tampered/real images to construct forensic VLMs. Beyond image-wise or pixel-wise classification, forensic VLMs take a suspected image and an instruction prompt as inputs, and output a probability of being fake, a natural-language justification, and a mask indicating the tampered region when the probability exceeds a threshold. These textual justifications provide auditable signals for downstream human-assisted review, even though their faithfulness can vary across models and deployment settings.

\begin{figure}[!t]
  \centering
  \includegraphics[width=0.95\linewidth]{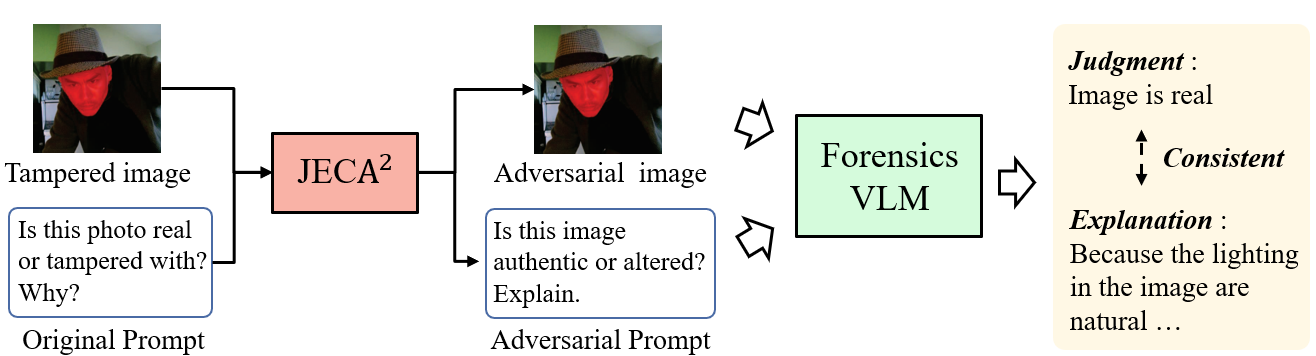}
  \caption{Schematic illustration of the proposed JECA$^2$ against a forensic VLM. The example visualizes the intended false-consistency outcome: the attacked model predicts ``Real'' and produces an explanation that supports this flipped judgment under the automated consistency metric.}
  \label{fig:teaser}
\end{figure}

Despite the advantage, it remains unclear whether forensic VLMs can effectively resist consistency-oriented adversarial stress tests. Recent studies \cite{b50,b36,b25} have shown that foundation VLMs are vulnerable to adversarial attacks. These attacks introduce crafted perturbations to the input image~\cite{b34,b10,b20}, the textual prompt~\cite{b25,b47}, or both~\cite{b50,b36}, thereby misleading the foundation VLMs into producing incorrect predictions. This raises a security-auditing concern that forensic VLMs fine-tuned from foundation models may also be susceptible to adversarial manipulation, especially when their outputs are used to triage cases for later human review.

Unfortunately, many existing adversarial attacks designed for foundation VLMs may be insufficient to expose the vulnerabilities of forensic VLMs. To mislead the VLMs, they push the perturbed visual or textual embedding away from its clean counterpart, or increase the cross-modal mismatch between image and prompt representations. While effective for degrading VLM predictions, most of them do not explicitly evaluate the coherence between the final decision and its accompanying explanation. As a result, a forensic VLM may judge the adversarially tampered images real but generate an explanation that still hints at manipulation cues. In a triage workflow, such inconsistency can serve as an audit signal for deciding whether a case needs escalation, even when the human reviewer later inspects the image itself. A recent concurrent anti-forensics study, ForgeryEraser~\cite{bForgeryEraser}, is closely related in showing that explainable forensic models can be induced to produce authenticity-consistent explanations for forged images; our work focuses on a judgment-explanation consistency formulation for forensic VLMs, together with attention-diversion, prompt-embedding optimization, and threat-level evaluation.

To address the above issue, we propose a Judgment-Explanation Consistent (JECA$^2$) adversarial attack against forensic VLMs. As shown in Fig.~\ref{fig:teaser}, on the visual side, JECA$^2$ crafts adversarial perturbations that redirect a Grad-CAM attribution proxy from manipulated regions toward non-manipulated areas, inducing a forensic VLM to output an incorrect decision. On the textual side, it optimizes the prompt embedding toward the target ``Real'' judgment under a token-proximity constraint and explicitly evaluates whether the resulting outputs remain judgment--explanation consistent. In the primary SID-Set/SIDA protocol, JECA$^2$ achieves higher attack success rate and automated judgment-explanation consistency than the implemented baselines.

To summarize, our contributions are threefold:
\begin{itemize}
    \item[1)] We provide a targeted study of adversarial vulnerability in image forensic VLMs, with particular focus on the consistency between judgment and explanation.

    \item[2)] We develop a novel Judgment--Explanation Consistent (JECA$^2$) adversarial attack, designed to induce forensic VLMs to produce incorrect yet internally consistent outputs under our automated consistency metric.
    
    \item[3)] Extensive experiments characterize the approach across datasets, threat conditions, predicted-mask variants, and transfer settings while explicitly separating the strongest diagnostic protocol from weaker deployment-like settings.
\end{itemize}

\section{Related Work}
\label{sec:related}

\subsection{Image Forensic}

Image forensics aims to determine the authenticity and integrity of visual content. Early methods rely on frequency-domain analysis~\cite{b2}, residual-based descriptors~\cite{b1}, and physiological cues such as eye blinking~\cite{b3}. With the advent of deep learning, convolutional neural networks have been widely adopted for binary classification (real vs.\ fake)~\cite{b85} and forgery localization~\cite{b60}. Despite their effectiveness, these methods provide only classification scores or segmentation masks without explaining why an image is deemed manipulated.

With the remarkable development of vision-language models (VLMs), recent works construct forensic VLMs that jointly perform detection, localization, and natural-language explanation. SIDA~\cite{b6} fine-tunes a large multimodal model on a massive social-media-scale dataset to simultaneously output a forgery probability, a textual explanation, and a tampering mask. FakeShield~\cite{b7} designs a multi-modal framework with a domain tag-guided detection module and a multi-modal forgery localization module for explainable image forgery detection and localization. ForgeryGPT~\cite{bForgeryGPT} captures high-order forensic knowledge correlations across diverse linguistic feature spaces, enabling interactive dialogue for forensic analysis. FFAA~\cite{bFFAA} integrates a fine-tuned multimodal LLM with a multi-answer intelligent decision system for open-world face forgery analysis. DF-LLaVA~\cite{bDFLLaVA} unlocks the intrinsic discrimination potential of MLLMs via knowledge injection and conflict-driven self-reflection, achieving both high accuracy and explainability. However, the adversarial robustness of these forensic VLMs remains largely unexplored, which motivates our work.

\subsection{Adversarial Attack on VLMs}

Based on the attack modality, existing adversarial attacks on VLMs can be categorized into unimodal and multimodal attacks.

\noindent\textit{Image-only attacks.} Classical methods such as FGSM~\cite{b34}, PGD~\cite{b10}, and C\&W~\cite{b20} craft pixel perturbations to fool classifiers. AnyAttack~\cite{b22} and PNA~\cite{b82} improve transferability across vision transformers. In the forensic domain, anti-forensic perturbations~\cite{b80,b33} suppress manipulation traces, and RLGC~\cite{bRLGC} uses reinforcement learning for black-box attacks on CNN-based detectors. However, these methods cannot address the textual reasoning pathway of VLM-based detectors.

\noindent\textit{Text-only attacks.} GCG~\cite{b25} and AutoDAN~\cite{b47} optimize adversarial prompts to jailbreak language models, but cannot alter the visual evidence that forensic systems attend to.

\noindent\textit{Cross-modal attacks.} CroPA~\cite{b50} demonstrates prompt-agnostic visual adversarial examples for VLMs. CMI~\cite{b36} improves transferability through collaborative multimodal interaction. JMTFA~\cite{b41} targets attention relevance scores, while MAA~\cite{b38} explores adversarial robustness of VLM grounding capabilities. Recent studies also show that visual adversarial signals can bypass task constraints in multimodal agents~\cite{bPennyWise} and poison multimodal memory for later planning~\cite{bVisualInception}, suggesting that VLM security risks extend beyond immediate label flipping. These methods optimize cross-modal feature alignment or downstream agent behavior but do not explicitly target the attention mechanism underpinning forensic reasoning.

Crucially, most of the above attacks do not evaluate the coherence between the final judgment and its accompanying explanation. As a result, a forensic VLM may be misled into predicting ``Real'' but still generate an explanation hinting at manipulation cues---an inconsistency that may be flagged during downstream explanation-based review. A concurrent work~\cite{bReasoningShifts} studies reasoning robustness of audio language models under adversarial attacks, but focuses on analysis rather than attack design and operates in the audio domain. ForgeryEraser~\cite{bForgeryEraser} is closely related in that it studies anti-forensics for explainable image forgery detection; JECA$^2$ focuses on forensic VLMs and explicitly optimizes judgment-explanation consistency through visual attention diversion and prompt-embedding alignment.


\section{The Proposed Method}
\label{sec:method}

Fig.~\ref{fig:workflow} illustrates the framework of our Judgment--Explanation Consistent Adversarial Attack (JECA$^2$), which consists of three modules:

\begin{figure*}[!t]
  \centering
  \includegraphics[width=0.9\linewidth]{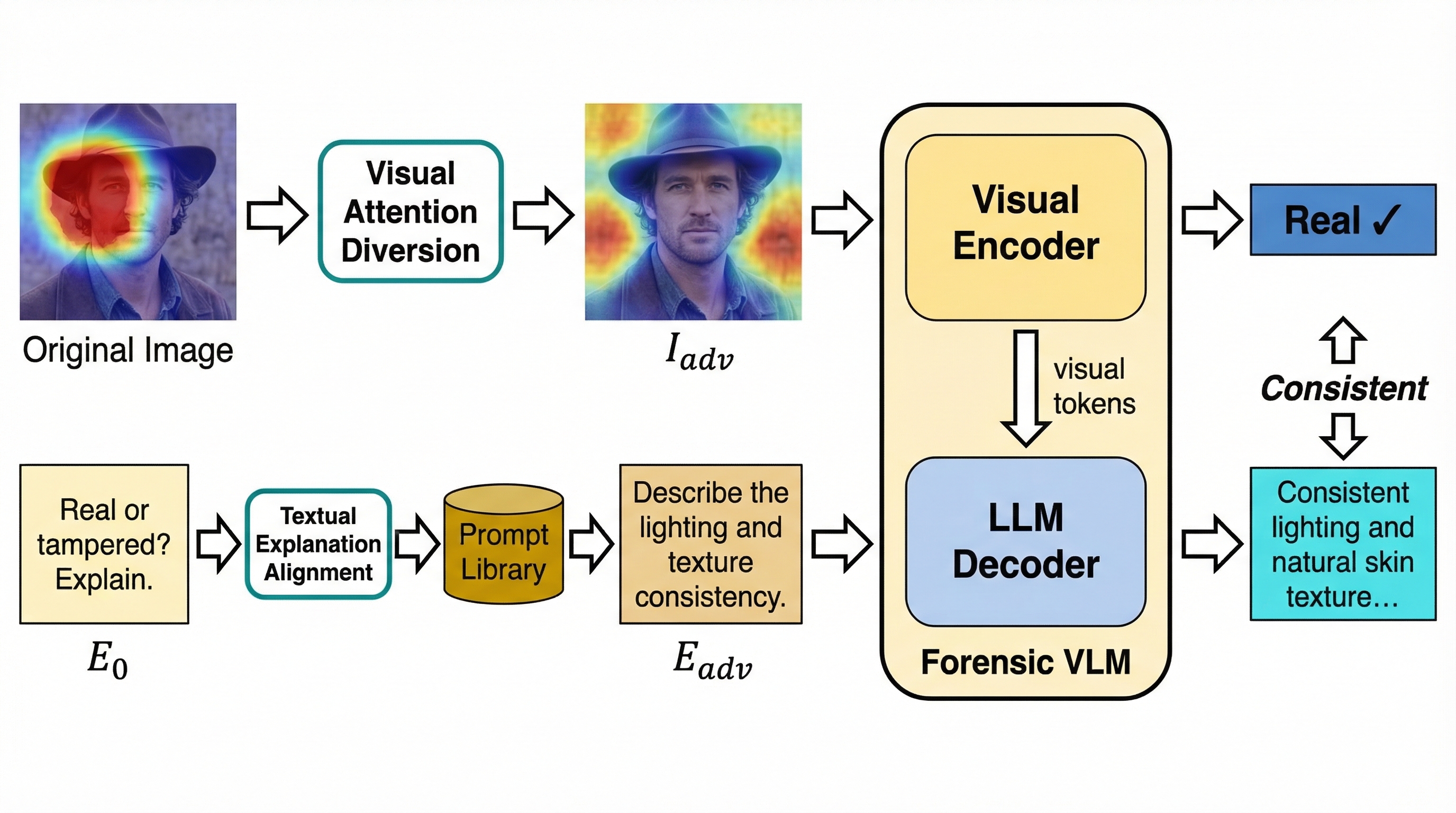}
  \caption{Complete workflow of the JECA$^2$ framework. The visual attention diversion module uses Grad-CAM and bidirectional attention interference to redirect attention from tampered regions ($R_{\text{tamper}}$) to background decoys ($R_{\text{bg}}$). The textual explanation alignment module optimizes prompt embeddings toward the target ``Real'' judgment under a token-proximity constraint; the ``Prompt Library'' box schematically denotes the vocabulary-anchor set used to keep optimized embeddings close to valid token semantics, rather than an additional discrete search module. Joint optimization alternates between the two modules for $T=100$ iterations, producing adversarial image $I_{\text{adv}}$ and optimized embedding $E_{\text{adv}}$.}
  \label{fig:workflow}
\end{figure*}
1) a visual attention diversion module that perturbs input images to redirect visual attention toward non-manipulated areas, inducing a forensic VLM to make an incorrect prediction; 2) a textual explanation alignment module that optimizes prompt embeddings under the stronger system-level threat model to encourage an explanation supporting the faulty judgment; and 3) a joint optimization module that alternately optimizes the two modalities for stable convergence in the controlled red-team protocol.

\subsection{Visual Attention Diversion}
\label{sec:vad}

To achieve judgment--explanation consistency, the first step is to redirect the forensic VLM's visual attribution proxy from manipulated regions toward non-manipulated areas. If the model relies more on benign regions, its flipped judgment and generated explanation are more likely to emphasize authentic-looking visual content, laying the foundation for false consistency.

Given an input image $I \in \mathbb{R}^{H \times W \times 3}$, perturbation $\delta$ ($\|\delta\|_\infty \le \epsilon$), and prompt embeddings $E \in \mathbb{R}^{n \times d}$, let $F(I, E)$ denote the VLM's prediction and $M(I; \delta) \in \mathbb{R}^{H \times W}$ denote the Grad-CAM~\cite{b35} attention map. We define $R_{\text{tamper}}$ as the set of tampered pixel coordinates and $R_{\text{bg}}$ as the background (non-tampered) region.

\textbf{Detection Loss.} The primary objective is to flip the model's classification:
\begin{equation}
\mathcal{L}_{\text{det}}(\delta) = \text{CE}(F_{\text{cls}}(I + \delta, E_{\text{fixed}}), y_{\text{real}})
\end{equation}

\textbf{Bidirectional Attention Interference.} We combine two complementary objectives:
\begin{equation}
\mathcal{L}_{\text{att}}(I) = \alpha \cdot \mathcal{L}_{\text{mislead}} + (1-\alpha) \cdot \mathcal{L}_{\text{hide}}
\label{eq:att}
\end{equation}
where $\mathcal{L}_{\text{mislead}}$ amplifies attention on decoy regions while suppressing attention on tampered regions, and $\mathcal{L}_{\text{hide}}$ smooths tampering boundaries via total variation regularization.

Specifically, we inject Gaussian pseudo-hotspots in $R_{\text{bg}}$ centered at a decoy location $(x_0, y_0)$ with spread $\sigma$ (selection details in supplementary material):
\begin{equation}
G_{i,j} = \exp\!\Bigl(-\tfrac{(i-x_0)^2+(j-y_0)^2}{2\sigma^2}\Bigr)
\end{equation}
The attention diversion loss combines decoy amplification with tampered-region suppression:
\begin{align}
\mathcal{L}_{\text{mislead}} = & - \frac{1}{|R_{\text{bg}}|} \sum_{(i,j) \in R_{\text{bg}}} M(I)_{i,j} \cdot G_{i,j} \nonumber \\
& + \lambda_s \cdot \frac{1}{|R_{\text{tamper}}|} \sum_{(i,j) \in R_{\text{tamper}}} M(I)_{i,j}
\end{align}
The concealment loss applies total variation on tampering boundaries:
\begin{equation}
\mathcal{L}_{\text{hide}} = \sum_{(i,j) \in R_{\text{tamper}}} \left( |(I{+}\delta)_{i+1,j} - (I{+}\delta)_{i,j}| + |(I{+}\delta)_{i,j+1} - (I{+}\delta)_{i,j}| \right)
\end{equation}

This bidirectional strategy is motivated by the relative competition induced by normalized self-attention: empirically, increasing saliency on a decoy region tends to reduce the relative attribution assigned to tampered regions. We treat Grad-CAM as an optimization proxy for model focus rather than as a faithful measurement of internal attention.

\subsection{Textual Explanation Alignment}
\label{sec:tea}

While visual attention diversion redirects the model's attribution proxy to benign regions, the LLM component of a forensic VLM may still generate explanations that hint at manipulation cues based on its prior knowledge. To reduce this gap, we optimize the prompt embeddings toward the ``Real'' decision while constraining them to remain close to valid token embeddings. We then assess authenticity-affirming explanations and spatial alignment empirically through the JEC and attribution analyses, matching the evaluation directly to the judgment--explanation consistency phenomenon studied in this work.

Concretely, we optimize prompt embeddings $E \in \mathbb{R}^{n \times d}$ to increase the model's confidence in the target ``Real'' judgment after the visual attack has changed the model's evidence distribution. The optimization objective is:
\begin{equation}
\mathcal{L}_{\text{semantic}}(E) = \text{CE}(F(I + \delta_{\text{fixed}}, E), y_{\text{real}}) + \beta \cdot \mathcal{L}_{\text{coherence}}
\end{equation}
where the cross-entropy term encourages the model to predict ``Real'' with high confidence, and $\mathcal{L}_{\text{coherence}}$ encourages the optimized embeddings to remain close to valid tokens:
\begin{equation}
\mathcal{L}_{\text{coherence}} = \sum_{i=1}^{n} \min_{v \in \mathcal{V}_k} \|e_i - v\|_2^2
\end{equation}
Here $\mathcal{V}_k$ denotes the $k$-nearest vocabulary embeddings ($k{=}100$, vocabulary size $|\mathcal{V}|{\approx}32$K), and $\beta{=}0.1$ balances classification and coherence objectives. The optimization trajectory is visualized in the supplementary material.

\textbf{Synergy with Visual Attention Diversion.} The two modules are complementary by design. The visual module encourages the model's Grad-CAM attribution proxy to shift toward non-manipulated areas; the textual module biases the optimized prompt representation toward the target ``Real'' decision under a token-proximity constraint. In successful cases, we empirically observe false consistency: the explanation references visual features (\eg, ``consistent lighting,'' ``natural skin texture'') while the attribution proxy emphasizes benign regions. This motivates our automated JEC evaluation and indicates a consistency failure mode that explanation-based triage systems should audit.

\textbf{Semantic Plausibility.} Nearest-token projections of the optimized embeddings show 94.2\% token validity and 89.1\% forensic term retention, and GPT-4 proxy scoring suggests plausible generated explanations (detailed protocol in supplementary material). More importantly, we evaluate judgment--explanation consistency across all attack methods in Sect.~\ref{sec:jec}, showing that JECA$^2$ achieves significantly higher consistency than baselines under the same automated proxy. This module requires embedding-level access (Threat Level~\Romannum{2}) and should be interpreted as a stronger diagnostic setting; under the same white-box SID-Set/SIDA attack protocol but without embedding modification, the visual attention diversion module alone achieves 80.3\% ASR.

\subsection{Joint Optimization}
\label{sec:joint}

Optimizing the visual perturbation $\delta$ and prompt embeddings $E$ simultaneously can cause instability due to their mutual dependence. We therefore employ alternating optimization: each iteration first updates $\delta$ via PGD (freezing $E$), then updates $E$ via gradient descent (freezing $\delta$). The respective loss functions are:
\begin{align}
\mathcal{L}_{\text{vis}} &= \mathcal{L}_{\text{det}}(\delta) + \lambda_{1}\mathcal{L}_{\text{att}} + \lambda_{2}\left\lVert\delta\right\rVert_{2}^{2} \label{eq:vis}\\
\mathcal{L}_{\text{text}} &= \mathcal{L}_{\text{semantic}}(E) \label{eq:text}
\end{align}

The complete procedure is summarized in Algorithm~\ref{alg:jeca}.

\begin{algorithm}[tb]
\caption{JECA$^2$ Joint Optimization}
\label{alg:jeca}
\begin{algorithmic}[1]
\REQUIRE Forged image $I$, tampering mask $M_{\text{mask}}$ (ground truth for oracle evaluation or estimated for the no-ground-truth-mask variant), initial prompt $E_0$
\ENSURE Adversarial image $I_{\text{adv}}$, optimized embedding $E_{\text{adv}}$
\STATE Initialize $\delta \leftarrow 0$, $E \leftarrow E_0$
\FOR{$t = 1$ to $T$}
    \STATE Compute attention map $M = \text{GradCAM}(F_{\text{vis}}(I+\delta))$
    \STATE \textit{// Visual Attention Diversion (freeze $E$, update $\delta$)}
    \STATE $\mathcal{L}_{\text{vis}} \leftarrow \mathcal{L}_{\text{det}}(\delta) + \lambda_1 \mathcal{L}_{\text{att}} + \lambda_2 \|\delta\|_2^2$
    \STATE $\delta \leftarrow \text{clip}(\delta - \eta_v \cdot \text{sign}(\nabla_\delta \mathcal{L}_{\text{vis}}),\; -\epsilon,\; \epsilon)$
    \STATE \textit{// Textual Explanation Alignment (freeze $\delta$, update $E$)}
    \STATE $\mathcal{L}_{\text{text}} \leftarrow \mathcal{L}_{\text{semantic}}(E)$
    \STATE $E \leftarrow E - \eta_e \cdot \nabla_E \mathcal{L}_{\text{text}}$
\ENDFOR
\RETURN $I_{\text{adv}} = I + \delta$, $E_{\text{adv}} = E$
\end{algorithmic}
\end{algorithm}

ASR stabilizes after $T{\approx}60$ iterations with diminishing returns thereafter; we use $T{=}100$ to ensure convergence across diverse image types.

\textbf{Hyperparameter Settings.} We set perturbation bound $\epsilon = 8/255$, attention balance $\alpha = 0.7$, coherence weight $\beta = 0.1$, loss weights $\lambda_1 = 1.0$, $\lambda_2 = 0.01$, suppression weight $\lambda_s = 1.0$, Gaussian spread $\sigma = 15$ pixels, learning rates $\eta_v = 1/255$ and $\eta_e = 0.01$, and $T = 100$ iterations. 
\subsubsection{Threat Model.}
\label{sec:threat}
We formalize the adversary's capabilities under two threat levels to separate visual-side robustness from stronger system-level access. \textbf{Level~\Romannum{1} (Image-Only Modification):} The adversary modifies only the input image under the same white-box optimization protocol and mask-access setting specified below; only the visual attention diversion module is applicable, achieving 80.3\% ASR on the SID-Set attack protocol described in Sect.~\ref{sec:exp}. \textbf{Level~\Romannum{2} (Image + Embedding Modification):} The adversary modifies both the input image and the prompt embedding, modeling compromised preprocessing, malicious deployment, or insider access to prompt representations rather than ordinary API-only use; the full JECA$^2$ achieves 87.2\% ASR under the same protocol. The 6.9\% gap quantifies the additional risk observed when embedding-level manipulation is possible. Our primary evaluation uses oracle tampering masks to isolate the judgment--explanation consistency vulnerability, while a DINO-saliency predicted-mask variant reaches 78.6\% ASR on the same 3,000-image test protocol without ground-truth masks. Thus, the oracle-mask setting is a diagnostic upper-bound protocol, and the predicted-mask variant characterizes a no-ground-truth-mask setting. Detailed Level~\Romannum{2} scenarios are provided in the supplementary material.

\section{Experiments}
\label{sec:exp}

\subsection{Experimental Setup}

We evaluate JECA$^2$ primarily on SID-Set~\cite{b6} using SIDA~\cite{b6} as the white-box target. We use OpenForensics~\cite{bOpenForensics} and FakeShield~\cite{b7} for transfer-style evaluations; when both dataset and model change, we explicitly interpret the result as a combined dataset/model transfer setting.

\textbf{Dataset Statistics.} Following the official SID-Set benchmark~\cite{b6}, the full dataset contains 300K images comprising 100K real images (from OpenImages V7), 100K fully synthetic images (generated by FLUX), and 100K tampered images (object replacement and partial manipulation). For the main SID-Set attack evaluation, we sample 3,000 tampered images from the official test split using the available manipulation metadata for stratification when provided. ASR is computed only on the subset initially classified as ``Fake,'' since otherwise baseline detector errors would be counted as attack success. OpenForensics~\cite{bOpenForensics} provides ${\sim}$115K images with pixel-level forgery masks; we sample 1,000 images per forgery type for the OpenForensics transfer-style evaluation.

\textbf{Evaluation Metrics.} We report Attack Success Rate (ASR), Joint ASR (J-ASR), localization IoU, and perceptual quality (PSNR, SSIM, LPIPS). ASR is computed as the fraction of forged images \textit{correctly detected in baseline} that are misclassified as ``Real'' after attack: $\text{ASR} = |\{x \in \mathcal{D}_{\text{det}} : F(x_{\text{adv}}) = \text{Real}\}| / |\mathcal{D}_{\text{det}}|$. This ensures ASR measures attack effectiveness rather than baseline model errors. J-ASR requires both classification flip \textit{and} IoU $<$ 0.2 (threshold justification in supplementary material). \textbf{Aux. Acc} is the auxiliary accuracy reported by the detector's evaluation script on the constructed attacked batch; because it uses a different denominator from ASR and depends on batch composition, we use ASR/J-ASR/IoU as the primary attack evidence. We also report paired significance diagnostics over the shared detectable subset.

\subsection{Comparison with Baseline Attacks}

\cref{tab:comparison} compares JECA$^2$ against baselines on the SID-Set attack subset. For JECA$^2$, we report mean$\pm$std over 5 random seeds; for re-implemented baselines, we report the deterministic run using the same sampled images and hyperparameter budget. All baselines are re-implemented under identical conditions ($\epsilon{=}8/255$, same target model and sampled images). Cross-modal methods (CroPA, CMI) are adapted from retrieval to detection objectives; these adapted baselines should be interpreted as fair in-protocol comparisons rather than exact reproductions of their original tasks.

\begin{table}[tb]
\caption{Attack performance of the proposed JECA$^2$ and the comparison methods on SID-Set. All baselines are re-implemented for fair comparison under identical conditions. J-ASR requires both classification flip and IoU$<$0.2. $p$-values are paired significance diagnostics vs.\ JECA$^2$ over per-image attack-success indicators on the shared detectable subset; we use them as supporting evidence rather than as the sole basis for the comparison.}
\label{tab:comparison}
\centering
\scriptsize
\setlength{\tabcolsep}{2.5pt}
\begin{tabular}{lcccccc}
\toprule
\textbf{Method} & \textbf{Aux. Acc}$\downarrow$ & \textbf{ASR}$\uparrow$ & \textbf{J-ASR}$\uparrow$ & \textbf{IoU}$\downarrow$ & \textbf{PSNR} & $p$\textbf{-value} \\
\midrule
No Attack & 93.5\% & -- & -- & 0.44 & -- & -- \\
FGSM~\cite{b34} & 59.3\% & 66.2\% & 54.1\% & 0.32 & 33.5 & $<$0.001 \\
PGD~\cite{b10} & 52.4\% & 72.8\% & 63.2\% & 0.27 & 36.2 & $<$0.001 \\
C\&W~\cite{b20} & 51.1\% & 73.5\% & 61.8\% & 0.28 & 39.1 & $<$0.001 \\
AnyAttack~\cite{b22} & 48.7\% & 74.9\% & 61.2\% & 0.25 & 36.2 & $<$0.001 \\
MAA~\cite{b38} & 46.2\% & 78.3\% & 67.5\% & 0.21 & 35.5 & $<$0.001 \\
JMTFA~\cite{b41} & 47.8\% & 76.1\% & 65.8\% & 0.22 & 36.1 & $<$0.001 \\
PNA~\cite{b82} & 48.5\% & 75.4\% & 64.2\% & 0.23 & 35.2 & $<$0.001 \\
\midrule
\multicolumn{7}{l}{\textit{Cross-modal/prompt-based baselines}} \\
CroPA~\cite{b50} & 45.2\% & 78.6\% & 69.1\% & 0.21 & 37.2 & $<$0.001 \\
CMI~\cite{b36} & 44.1\% & 79.5\% & 70.2\% & 0.20 & 36.4 & $<$0.01 \\
\midrule
\multicolumn{7}{l}{\textit{Task-specific black-box}} \\
RLGC~\cite{bRLGC} & 50.2\% & 74.3\% & 63.8\% & 0.24 & 35.8 & $<$0.001 \\
\midrule
\textbf{JECA$^2$} & \textbf{39.5$\pm$2.8\%} & \textbf{87.2$\pm$2.4\%} & \textbf{82.6$\pm$2.7\%} & \textbf{0.13$\pm$0.04} & 35.1 & -- \\
\bottomrule
\end{tabular}
\end{table}

JECA$^2$ outperforms all implemented baselines in this protocol. The 7.7\% improvement over the strongest baseline CMI (87.2\% vs 79.5\%) is consistent with the effect of directly manipulating the attribution proxy via $\mathcal{L}_{\text{mislead}}$, rather than only optimizing cross-modal feature alignment. Perceptual quality remains reasonable (SSIM=0.941, LPIPS=0.042), while localization degrades from IoU=0.44 to IoU=0.13 (70\% reduction).

\textbf{Attention Diversion Score (ADS).} We introduce ADS to quantify attention shift:
\begin{equation}
\text{ADS} = \frac{\sum_{(i,j) \in R_{\text{bg}}} M(I_{\text{adv}})_{i,j}}{\sum_{(i,j) \in R_{\text{bg}}} M(I_{\text{adv}})_{i,j} + \sum_{(i,j) \in R_{\text{tamper}}} M(I_{\text{adv}})_{i,j}}
\end{equation}
ADS measures the fraction of attribution-proxy mass allocated to background regions. Clean images yield ADS$\approx$0.12; JECA$^2$ amplifies this to ADS$\geq$0.7, shifting the proxy majority away from tampered regions.

\subsection{Qualitative Results and Explanation Consistency}

\begin{figure*}[tb]
  \centering
  \includegraphics[width=0.78\linewidth]{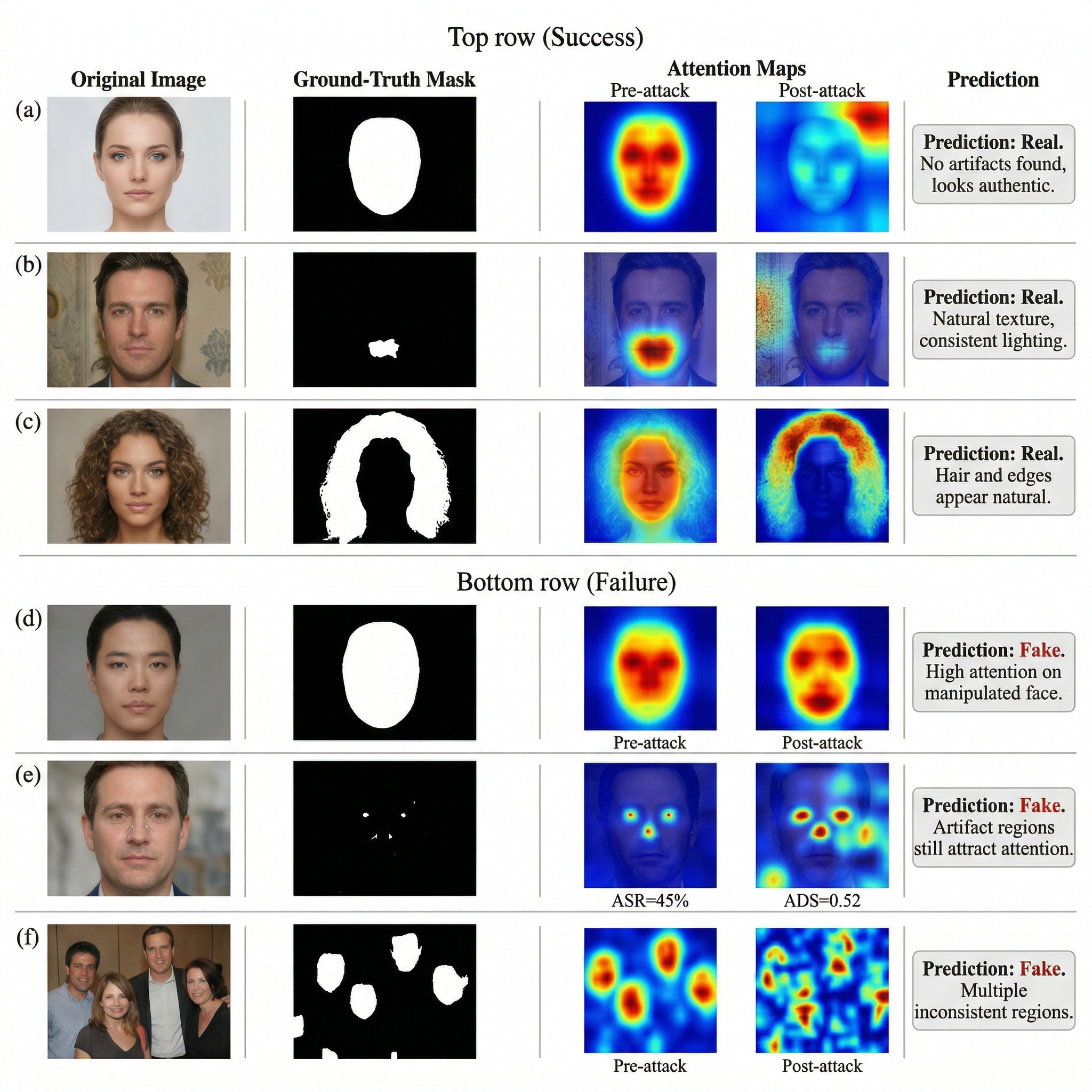}
  \caption{Success and failure cases of JECA$^2$. \textbf{Top (Success)}: (a) Face-swap; (b) Inpainting; (c) Hair/boundary manipulation. \textbf{Bottom (Failure)}: (d) Full-face manipulation; (e) High-frequency artifacts (ADS=0.52); (f) Multi-region tampering.}
  \label{fig:cases}
\end{figure*}

\textbf{Success cases (a--c) in \cref{fig:cases}.} JECA$^2$ succeeds when sufficient non-manipulated background exists to serve as an attention decoy. In face-swap (a), inpainting (b), and hair/boundary manipulation (c), the background occupies a large portion of the frame, enabling the Gaussian pseudo-hotspot in $R_{\text{bg}}$ to attract strong alternative attention and suppress forensic signals on $R_{\text{tamper}}$.

\textbf{Failure analysis.} We observe degraded performance under three conditions. \textit{(d) Full-face manipulation:} When the manipulated region $R_{\text{tamper}}$ covers nearly the entire face, $R_{\text{bg}}$ is too sparse to host a meaningful Gaussian decoy, so $\mathcal{L}_{\text{mislead}}$ has limited room to amplify alternative attribution and the visual attack often stalls. \textit{(e) High-frequency artifacts:} GAN compression artifacts can activate spatially diffuse attribution hotspots across the entire image; the total-variation component $\mathcal{L}_{\text{hide}}$ targets only $R_{\text{tamper}}$ boundaries and may not suppress these globally distributed signals, limiting ADS to 0.52 and leaving the model's forensic priors partially intact. \textit{(f) Multi-region tampering:} Multiple disjoint $R_{\text{tamper}}$ regions compete for the bounded perturbation budget ($\|\delta\|_\infty \le \epsilon$), while the single Gaussian decoy can redirect attribution toward only one background cluster at a time; without per-region decoys, overall ADS often remains insufficient to flip the model's judgment.

\textbf{Judgment--Explanation Consistency (JEC).}
\label{sec:jec}
To quantify whether explanations support the flipped judgment, we use GPT-4 as an automated proxy evaluator to rate consistency (1--5) on up to 500 stratified samples and report scores only on successfully attacked cases, since consistency is meaningful after the judgment has flipped. As shown in \cref{tab:jec}, image-only attacks achieve low JEC (2.05--2.31) as explanations still reference manipulation cues; cross-modal attacks achieve moderate JEC (2.85--3.12); JECA$^2$ achieves 4.15 with 87.6\% scoring $\geq$4. These results show that JECA$^2$ yields higher automated judgment-explanation consistency among successful attacks under a shared evaluator.

Because \cref{tab:jec} is conditional on each method's successful attacks, $N_{\text{eval}}$ varies with ASR and the table should be read as an automated consistency proxy rather than a human-validated forensic judgment. To make this analysis auditable, the supplementary material specifies fixed-$N$ and common-success subset protocols, together with a human-rating protocol for validating GPT-4 JEC scores when annotation resources are available.

\begin{table}[tb]
\caption{Judgment--Explanation Consistency (JEC) on SID-Set, conditional on successful attacks. JEC is rated 1--5 by GPT-4 as an automated proxy on successfully attacked samples; higher is more consistent. JECA$^2$ achieves JEC=4.15 with 87.6\% of successful samples scoring $\geq$4, substantially outperforming image-only attacks (JEC$\leq$2.31) and cross-modal attacks (JEC$\leq$3.12) under this conditional proxy metric. Equalized-subset and human-validation protocols are provided in the supplementary material to avoid over-interpreting the varying $N_{\text{eval}}$.}
\label{tab:jec}
\centering
\small
\begin{tabular}{llccc}
\toprule
\textbf{Type} & \textbf{Method} & $N_{\text{eval}}$ & \textbf{JEC}$\uparrow$ & \textbf{JEC$\geq$4 (\%)} \\
\midrule
\multirow{2}{*}{Image-only} & FGSM & 331 & 2.05 & 11.2\% \\
& PGD & 364 & 2.31 & 14.6\% \\
\midrule
\multirow{2}{*}{Cross-modal} & CroPA & 393 & 2.85 & 28.5\% \\
& CMI & 398 & 3.12 & 35.2\% \\
\midrule
Ours & \textbf{JECA$^2$} & 436 & \textbf{4.15} & \textbf{87.6\%} \\
\bottomrule
\multicolumn{5}{l}{\footnotesize $^\dagger$Evaluated on successfully attacked samples only ($N_{\text{eval}}$ varies by method's ASR).}
\end{tabular}
\end{table}

The large JEC gap---image-only attacks score only 2.05--2.31 while JECA$^2$ achieves 4.15---suggests a limitation of visual-only adversarial strategies under our proxy evaluator: even when the classification is successfully flipped, the LLM component can retain forensic language priors that surface in the generated explanation (\eg, referencing ``boundary artifacts'' or ``inconsistent textures''). Cross-modal baselines partially address this through feature-space alignment, achieving JEC up to 3.12, but without explicit target-judgment embedding optimization the LLM's prior can still leak manipulation cues into the output. The textual explanation alignment module reduces this gap by steering the optimized embedding $E_{\text{adv}}$ toward the target ``Real'' decision while preserving token proximity, providing one explanation for the step-wise JEC improvement observed across method classes under the automated metric. 
\subsection{Cross-Dataset Evaluation}

\cref{tab:ff} presents cross-dataset evaluation on OpenForensics using the full JECA$^2$ framework (Level~\Romannum{2}).

\begin{table}[tb]
\caption{OpenForensics transfer-style evaluation (full JECA$^2$ framework). JECA$^2$ achieves average ASR=77.8\% and J-ASR=71.2\% across four forgery types, with localization IoU dropping to 0.18 on average. This setting jointly reflects dataset and target-model transfer beyond the primary SID-Set/SIDA protocol.}
\label{tab:ff}
\centering
\small
\begin{tabular}{lcccc}
\toprule
\textbf{Forgery Type} & \textbf{ASR}$\uparrow$ & \textbf{J-ASR}$\uparrow$ & \textbf{IoU}$\downarrow$ & \textbf{PSNR} \\
\midrule
Face-swap & 80.2\% & 72.1\% & 0.17 & 35.8 \\
Inpainting & 76.2\% & 71.5\% & 0.18 & 36.1 \\
Splicing & 79.5\% & 73.1\% & 0.17 & 35.9 \\
Copy-move & 75.4\% & 67.9\% & 0.20 & 36.3 \\
\midrule
\textbf{Average} & \textbf{77.8\%} & \textbf{71.2\%} & \textbf{0.18} & 36.0 \\
\bottomrule
\end{tabular}
\end{table}

Across forgery types, face-swap (ASR=80.2\%) and splicing (79.5\%) consistently outperform copy-move (75.4\%) because their naturally larger non-manipulated background areas provide richer $R_{\text{bg}}$ for Gaussian pseudo-hotspot injection. Copy-move exhibits the lowest J-ASR (67.9\%) because copied regions are semantically similar to their source areas, causing the model to partially sustain forensic attention on multiple locations simultaneously. The 9.4\% ASR drop from the primary SID-Set/SIDA protocol (87.2\%) to the OpenForensics/FakeShield transfer-style protocol (77.8\%) should be interpreted as a combined dataset/model shift rather than as a pure dataset-transfer measurement.

\subsection{Ablation Study}

\begin{table}[tb]
\caption{Ablation Study on JECA$^2$ Components (SID-Set). L-IoUR (Localization IoU Reduction) measures the relative reduction in IoU: $1 - \text{IoU}_{\text{adv}}/\text{IoU}_{\text{clean}}$.}
\label{tab:ablation}
\centering
\small
\begin{tabular}{lccccc}
\toprule
\textbf{Config.} & \textbf{Vis.} & \textbf{Text.} & \textbf{Aux. Acc}$\downarrow$ & \textbf{ASR}$\uparrow$ & \textbf{L-IoUR}$\uparrow$ \\
\midrule
Vis. only & $\checkmark$ & $\times$ & 45.6\% & 80.3\% & 0.62 \\
Text. only & $\times$ & $\checkmark$ & 52.8\% & 73.5\% & 0.18 \\
w/o $\mathcal{L}_{\text{mislead}}$ & $\checkmark$ & $\checkmark$ & 44.2\% & 79.6\% & 0.48 \\
w/o $\mathcal{L}_{\text{hide}}$ & $\checkmark$ & $\checkmark$ & 40.8\% & 85.9\% & 0.68 \\
Full JECA$^2$ & $\checkmark$ & $\checkmark$ & \textbf{39.5\%} & \textbf{87.2\%} & \textbf{0.72} \\
\bottomrule
\end{tabular}
\end{table}

\textbf{Analysis.} Under Level~\Romannum{1} (image-only modification within the same white-box SID-Set/SIDA protocol), the visual module alone achieves 80.3\% ASR---competitive with the strongest cross-modal baseline CMI (79.5\%). Adding the textual module (Level~\Romannum{2}) yields 87.2\% ASR (+6.9\%), quantifying the additional risk observed in our controlled protocol when embedding modification is allowed. The textual module alone achieves 73.5\% ASR but poor localization disruption (L-IoUR=0.18), suggesting that visual attention diversion is the main driver of localization failure. L-IoUR improves from 0.62 (visual only) to 0.72 (full), a 16\% gain from the complementary synergy. The predicted-mask/no-ground-truth variant reaches 78.6\% ASR on the same 3,000-image protocol, indicating that the effect is not solely an artifact of oracle masks. JECA$^2$ is robust to the tested hyperparameter ranges (ASR varies $<$3\% for $\alpha$ and $\sigma$) and requires 8.2s per image on RTX 4090 (detailed analysis in supplementary material).

\subsection{Transferability Analysis}

Transfer attacks from SIDA to FakeShield achieve 38.5\% ASR (45.2\% with ensemble surrogates), possibly benefiting from shared ViT-based visual encoders. \cref{tab:blackbox} reports a snapshot evaluation on closed-source general-purpose VLMs, where transfer reaches 11.9--18.2\% ASR on each model's detectable subset (compared to 2.7\% for random noise). These results complement the controlled forensic-VLM evaluation but also show that transfer to fully opaque services is limited; we therefore interpret the closed-source results as measurable attenuation rather than as evidence of a strong practical black-box attack.

\textbf{Architecture-Dependent Transferability.} The effectiveness of JECA$^2$'s visual attention diversion is closely tied to the target model's attention mechanism. Cross-architecture analysis (detailed in supplementary material) shows that ViT variants exhibit the highest white-box ASR ($>$85\%). One plausible explanation is that normalized self-attention creates stronger relative competition between decoy and artifact regions, making the Grad-CAM proxy more responsive to decoy amplification. Swin Transformer achieves intermediate performance (ASR=84.7\%) because its windowed attention localizes this competition within windows. ConvNeXt shows moderate degradation (ASR=83.4\%) as it lacks explicit self-attention; Grad-CAM captures gradient-weighted feature importance rather than attention, making the diversion mechanism less precise. These architecture-specific observations partially explain the transfer gap to closed-source VLMs, whose visual backbones and preprocessing pipelines are not fully known.

\textbf{Improving Transferability via Ensemble Surrogates.} The 45.2\% ASR achieved through ensemble surrogate training (vs.\ 38.5\% single-source) shows that, in our protocol, optimizing across multiple surrogate architectures improves transferability within the evaluated forensic-model family. We construct the ensemble by jointly optimizing adversarial perturbations against SIDA (ViT-L/14) and FakeShield (ViT-B/16) with equal loss weighting, encouraging perturbations that exploit shared forensic representations rather than architecture-specific vulnerabilities. The 6.7\% improvement is consistent with the presence of shared artifact-detection patterns across the evaluated forensic VLMs, but it does not close the gap to fully opaque black-box services.

\begin{table}[tb]
\caption{Black-Box Transfer to Closed-Source VLMs. Baseline Acc: accuracy on clean forged images (binary Real/Fake). ASR is computed on the detectable subset $\mathcal{D}_{\text{det}}$ per our definition. Model names denote API families; the supplementary material provides the snapshot evaluation protocol.}
\label{tab:blackbox}
\centering
\small
\begin{tabular}{lcccc}
\toprule
\textbf{Target Model} & \textbf{Baseline Acc} & $|\mathcal{D}_{\text{det}}|$ & \textbf{ASR}$\uparrow$ & \textbf{95\% CI} \\
\midrule
GPT-4V & 78.7\% & 2,361 & 14.6\% & [12.8, 16.4] \\
Claude-3 Opus & 81.3\% & 2,439 & 11.9\% & [10.2, 13.6] \\
Gemini Pro & 75.5\% & 2,265 & 18.2\% & [16.1, 20.3] \\
Random noise$^\dagger$ & -- & -- & 2.7\% & [1.8, 3.6] \\
\bottomrule
\multicolumn{5}{l}{\scriptsize $^\dagger$Random noise ($\epsilon=8/255$): average ASR across all models.}
\end{tabular}
\end{table}

\subsection{Defense Evaluation}

\cref{tab:defense} summarizes defense effectiveness under white-box settings on SIDA. Unless otherwise stated, these are non-adaptive evaluations: the attack is generated without differentiating through the evaluated defense, and the defense is then applied at inference time. This protocol provides a controlled robustness diagnostic for common defense families.

\begin{table}[tb]
\caption{Defense Evaluation Under White-Box Settings on SIDA. Combined$^\dagger$ applies DiffPure preprocessing followed by MMCoA-enhanced inference.}
\label{tab:defense}
\centering
\small
\begin{tabular}{lccc}
\toprule
\textbf{Defense} & \textbf{ASR}$\downarrow$ & \textbf{IoU}$\uparrow$ & \textbf{Overhead} \\
\midrule
No Defense & 87.2\% & 0.13 & 1$\times$ \\
Adversarial Training & 66.8\% & 0.24 & 1.1$\times$ \\
Attention Entropy & 73.5\% & 0.21 & 1.05$\times$ \\
DiffPure~\cite{b63} & 63.2\% & 0.28 & 15$\times$ \\
MMCoA~\cite{b52} & 60.4\% & 0.27 & 1.2$\times$ \\
Combined$^\dagger$ & 47.8\% & 0.32 & 16$\times$ \\
\bottomrule
\end{tabular}
\end{table}

MMCoA~\cite{b52} provides the best single-defense trade-off in our non-adaptive protocol (60.4\% ASR at 1.2$\times$ overhead) via cross-modal consistency verification. The combined defense (DiffPure~\cite{b63} + MMCoA) achieves 47.8\% ASR but at 16$\times$ overhead, illustrating a robustness-efficiency trade-off for the evaluated defenses. Notably, attention entropy regularization achieves only 73.5\% ASR, suggesting that JECA$^2$'s pseudo-hotspot strategy is not fully captured by this entropy-based detector in the tested setting.

\subsection{Discussion and Limitations}

Our results suggest that VLM forensic systems' semantic consistency mechanism can become an attack surface~\cite{bFViT,bAdvXAI,bXAIDetAdv}. The strongest protocol reaches 87.2\% ASR by isolating the vulnerability under controlled white-box access, while the image-only and predicted-mask variants still achieve 80.3\% and 78.6\% ASR, respectively. This pattern indicates that the diagnostic signal is not limited to embedding manipulation or ground-truth mask access, although these stronger assumptions should be read as upper-bound stress tests rather than ordinary deployment conditions.

\textbf{Security Implications for Deployment.} The high ASR achieved by JECA$^2$ in the strongest setting raises concerns for forensic pipelines whose model weights, prompts, or preprocessing steps are accessible to attackers. In particular, forensic VLMs deployed with fixed and publicly documented architectures may expose more information than fully opaque services, but our closed-source snapshot shows only limited transfer. The 6.9\% ASR gap between Level~\Romannum{1} (image-only modification, 80.3\%) and Level~\Romannum{2} (image + embedding modification, 87.2\%) quantifies the marginal risk observed in our controlled protocol when embedding-level modification is allowed. This suggests that system designers should consider prompt integrity verification---\eg, cryptographic signing of prompt templates---as one low-cost countermeasure. Furthermore, the success of our attention diversion strategy without embedding modification (Level~\Romannum{1}) indicates that systems with protected prompt interfaces can still require visual-side robustness evaluation.

\textbf{Failure Mode Analysis.} The qualitative cases in \cref{fig:cases} show that performance decreases when decoy placement is constrained, high-frequency artifacts are spatially diffuse, or multiple tampered regions compete for the single-decoy budget. These cases point to natural extensions such as multi-hotspot injection, adaptive decoy allocation, and frequency-aware perturbation design.

\textbf{Future Defense Directions.} Our non-adaptive defense evaluation (\cref{tab:defense}) shows that no single tested defense reduces ASR below 60\% without substantial computational overhead. We identify three defense directions suggested by the failure modes: (1) \textit{attention consistency verification}, which cross-checks attribution maps across multiple inference passes with stochastic input augmentations to detect artificially concentrated patterns; (2) \textit{explanation grounding}, which verifies that generated explanations reference spatial regions consistent with attribution or localization evidence, detecting the false consistency that JECA$^2$ exploits; and (3) \textit{prompt embedding anomaly detection}, which monitors embedding-space trajectories for out-of-distribution patterns indicative of adversarial optimization.

\textbf{Scope and Limitations.} JECA$^2$ is evaluated most extensively under white-box access, with oracle masks used in the primary protocol to cleanly study judgment--explanation consistency. The predicted-mask, image-only, transfer, and defense studies broaden the evidence beyond this strongest setting, while fully opaque deployments, human-validated consistency ratings, and fully adaptive defense evaluations remain important follow-up settings.

\section{Conclusion}
\label{sec:conclusion}

We presented JECA$^2$, an adversarial framework for studying consistency between judgments and explanations in forensic VLMs under explicit threat models. By jointly attacking both the visual and textual pathways---redirecting an attribution proxy to benign regions while optimizing prompt embeddings toward the target judgment under a token-proximity constraint---JECA$^2$ induces forensic VLMs to produce incorrect yet internally consistent outputs under our automated consistency metric. The visual attention diversion module alone achieves 80.3\% ASR under image-only modification; the full diagnostic framework reaches 87.2\% ASR with joint optimization; and a predicted-mask variant reaches 78.6\% ASR without ground-truth masks. Together with the limited closed-source transfer results, these findings highlight prompt embedding integrity and visual-side attribution robustness as important considerations for forensic VLM red-team evaluation.

\noindent\textbf{Ethics Statement.} We acknowledge the dual-use nature of this research and will follow responsible disclosure before public release. Attack code will only be shared with verified researchers. Full ethics discussion is provided in the supplementary material.

\clearpage
\section*{Supplementary Material}
\setcounter{section}{0}
\renewcommand{\thesection}{A\arabic{section}}
\renewcommand{\theHsection}{supp.\arabic{section}}
\renewcommand{\theHsubsection}{supp.\arabic{section}.\arabic{subsection}}
\setcounter{figure}{0}
\renewcommand{\thefigure}{A\arabic{figure}}
\renewcommand{\theHfigure}{supp.\arabic{figure}}
\setcounter{table}{0}
\renewcommand{\thetable}{A\arabic{table}}
\renewcommand{\theHtable}{supp.\arabic{table}}
\setcounter{algorithm}{0}
\renewcommand{\thealgorithm}{A\arabic{algorithm}}
\renewcommand{\theHalgorithm}{supp.\arabic{algorithm}}
\setcounter{equation}{0}
\renewcommand{\theequation}{A\arabic{equation}}
\renewcommand{\theHequation}{supp.\arabic{equation}}
\input{supplementary}

\bibliographystyle{splncs04}
\bibliography{main}
\end{document}

%% file: supplementary.tex

\section{Complete Algorithm Pseudocode}
\label{sec:supp_algorithm}

\begin{algorithm}[tb]
\caption{JECA$^2$ Alternating Optimization}
\label{alg:camap}
\begin{algorithmic}[1]
\FOR{$t = 1$ to $T$}
    \STATE Compute Attention Map $M = \text{GradCAM}(F_{\text{vis}}(I+\delta))$
    \STATE Compute Prediction $\hat{y} = F(I+\delta, E)$
    \STATE \textbf{Loss Calculation:}
    \STATE $\mathcal{L}_{\text{mislead}} \leftarrow -\frac{1}{|R_{\text{bg}}|}\sum_{(i,j) \in R_{\text{bg}}} M(I+\delta)_{i,j} \cdot G_{i,j} + \frac{\lambda_s}{|R_{\text{tamper}}|} \sum_{(i,j) \in R_{\text{tamper}}} M(I+\delta)_{i,j}$
    \STATE $\mathcal{L}_{\text{hide}} \leftarrow \text{TV}(I+\delta, M_{\text{tamper}})$
    \STATE $\mathcal{L}_{\text{att}} \leftarrow \alpha \cdot \mathcal{L}_{\text{mislead}} + (1-\alpha) \cdot \mathcal{L}_{\text{hide}}$ (Eq.~(2) in main paper)
    \STATE $\mathcal{L}_{\text{det}}(\delta) = \text{CE}(\hat{y}, y_{\text{real}})$ with $E$ frozen (Eq.~(1) in main paper)
    \STATE $\mathcal{L}_{\text{vis}} = \mathcal{L}_{\text{det}}(\delta) + \lambda_1 \mathcal{L}_{\text{att}} + \lambda_2 \|\delta\|_2^2$ (Eq.~(6) in main paper)
    \STATE \textbf{Visual Update:} \textit{// Freeze $E$, update $\delta$ only}
    \STATE $\delta \leftarrow \delta - \eta_v \cdot \text{sign}(\nabla_\delta \mathcal{L}_{\text{vis}})$
    \STATE $\delta \leftarrow \text{clip}(\delta, -\epsilon, \epsilon)$ \textit{// PGD projection}
    \STATE \textbf{Text Update:} \textit{// Freeze $\delta$, update $E$ only}
    \STATE $\mathcal{L}_{\text{semantic}}(E) = \text{CE}(F(I+\delta, E), y_{\text{real}}) + \beta \cdot \mathcal{L}_{\text{coherence}}$ (Eq.~(5) in main paper)
    \STATE $E \leftarrow E - \eta_e \cdot \nabla_E \mathcal{L}_{\text{semantic}}$
\ENDFOR
\STATE $I_{\text{adv}} \leftarrow I + \delta$, $E_{\text{adv}} \leftarrow E$
\RETURN $I_{\text{adv}}$, $E_{\text{adv}}$
\end{algorithmic}
\end{algorithm}

\section{Attention Aggregation Details}
\label{sec:supp_attention}

This section provides additional technical details on the attention map aggregation procedure used by the visual attention diversion module (Sect.~3.1 of the main paper).

For multi-head self-attention in ViT-L/14 (16 heads per layer), we aggregate attention maps via gradient-weighted averaging across heads within each layer, then uniform averaging across layers 18--23. This yields a single spatial attention map $M \in \mathbb{R}^{H \times W}$ representing the model's focus distribution.

\textbf{Layer Selection Rationale.} We select layers 18--23 (out of 24 total layers) based on empirical correlation analysis with SIDA's output segmentation masks. Earlier layers (0--12) capture low-level features with weak localization correlation ($r < 0.5$); middle layers (13--17) show moderate correlation ($r \in [0.5, 0.65]$); later layers (18--23) achieve highest correlation ($r \in [0.71, 0.75]$) as they encode semantically meaningful patterns relevant to forgery detection.

Specifically, for each layer $l \in \{18, 19, \ldots, 23\}$ and head $h \in \{1, 2, \ldots, 16\}$, we compute the layer-wise aggregated attention:
\begin{equation}
M_l = \sum_{h=1}^{16} w_{l,h} \cdot A_{l,h}
\end{equation}
where $A_{l,h} \in \mathbb{R}^{H \times W}$ is the attention map from head $h$ in layer $l$, and $w_{l,h}$ is the gradient weight computed as:
\begin{equation}
w_{l,h} = \text{ReLU}\left(\frac{1}{HW}\sum_{i,j} \frac{\partial \mathcal{L}_{\text{det}}}{\partial A_{l,h}^{i,j}}\right)
\end{equation}
The ReLU ensures only positive contributions are considered, following the standard Grad-CAM formulation~\cite{b35}.

The final aggregated attention map is obtained by uniform averaging across selected layers:
\begin{equation}
M = \frac{1}{6} \sum_{l=18}^{23} M_l
\end{equation}

This two-stage aggregation (gradient-weighted within layers, uniform across layers) balances task-relevance with robustness to layer-specific noise.

\subsection{Grad-CAM Differentiability and Optimization Stability}
\label{sec:supp_gradcam_stability}

Optimizing through Grad-CAM introduces second-order dependencies that can cause gradient instability. Specifically, Grad-CAM computes:
\begin{equation}
M = \text{ReLU}\left(\sum_k \alpha_k A^k\right), \quad \alpha_k = \frac{1}{Z}\sum_{i,j} \text{ReLU}\left(\frac{\partial y^c}{\partial A^k_{ij}}\right)
\end{equation}
where $A^k$ are feature maps, $\alpha_k$ are gradient-weighted importance scores, and $Z = H \times W$ is the spatial normalization factor. Differentiating through this expression yields second-order terms $\frac{\partial^2 y^c}{\partial A^k \partial \delta}$ that can exhibit high variance.

\textbf{Mitigation Strategies:} We employ three techniques to stabilize optimization: gradient clipping (max norm at 1.0) prevents gradient explosion from second-order terms; linear learning rate warmup over the first 10 iterations allows gradients to stabilize before aggressive updates; and exponential moving average of attention maps ($\gamma=0.9$) smooths optimization targets via $\bar{M}_t = \gamma \bar{M}_{t-1} + (1-\gamma) M_t$.

\textbf{Empirical Analysis:} Table~\ref{tab:stability} summarizes the impact of each mitigation strategy.

\begin{table}[tb]
\caption{Impact of Stability Mitigations on Optimization.}
\label{tab:stability}
\centering
\small
\begin{tabular}{lccc}
\toprule
\textbf{Configuration} & \textbf{Divergence Rate} & \textbf{Final ASR} & \textbf{Iterations} \\
\midrule
No mitigations & 23.4\% & 71.2\% & 100 \\
+ Gradient clipping & 8.5\% & 82.5\% & 100 \\
+ LR warmup & 3.2\% & 86.8\% & 100 \\
+ EMA (full) & 0.8\% & 87.2\% & 100 \\
\bottomrule
\end{tabular}
\end{table}

Without these mitigations, optimization diverges in 23.4\% of cases (loss $>$ 100 after 50 iterations). With all mitigations, divergence rate drops to $<$1\%, with mean loss reduction of 85\% over 100 iterations. The EMA smoothing is particularly important for preventing oscillation in attention map targets.

\section{Decoy Region Selection Strategy}
\label{sec:supp_decoy}

This section details the decoy region selection procedure for the visual attention diversion module, corresponding to the Gaussian pseudo-hotspot mechanism in Sect.~3.1 of the main paper (Eq.~(3)).

We employ a three-stage selection process for the benign background region $R_{\text{bg}}$:

\textbf{Stage 1: Semantic Filtering.} We exclude regions containing faces, text, or salient objects using DINO~\cite{b96} features. Specifically, we compute the self-attention maps from DINO-ViT and threshold at 0.5 to identify salient regions.

\textbf{Stage 2: Texture Analysis.} We prioritize regions with moderate texture complexity (entropy $\in [4.0, 6.0]$ bits) to ensure perturbations appear natural. Texture entropy is computed over $32 \times 32$ patches using:
\begin{equation}
H = -\sum_{i} p_i \log_2 p_i
\end{equation}
where $p_i$ is the normalized histogram of gradient magnitudes.

\textbf{Stage 3: Spatial Constraints.} We select regions at least 50 pixels from image boundaries and $>$100 pixels from $R_{\text{tamper}}$ to avoid overlap.

For each image, we sample $K=3$ candidate centers $(x_0, y_0)$ satisfying these criteria and select the one maximizing initial attention response $\sum_{i \in R_{\text{bg}}} M(I)_i$.

\subsection{Impact of Decoy Selection on Attack Performance}

Table~\ref{tab:decoy_ablation} analyzes the sensitivity of JECA$^2$ to decoy region selection parameters.

\begin{table}[tb]
\caption{Sensitivity Analysis of Decoy Selection Parameters on SID-Set.}
\label{tab:decoy_ablation}
\centering
\small
\begin{tabular}{lccc}
\toprule
\textbf{Configuration} & \textbf{ASR}$\uparrow$ & \textbf{ADS}$\uparrow$ & \textbf{PSNR} \\
\midrule
\multicolumn{4}{l}{\textit{Gaussian spread $\sigma$ (pixels)}} \\
$\sigma = 5$ & 82.0\% & 0.66 & 36.1 \\
$\sigma = 10$ & 85.3\% & 0.72 & 35.6 \\
$\sigma = 15$ (default) & 87.2\% & 0.76 & 35.1 \\
$\sigma = 20$ & 85.9\% & 0.74 & 34.5 \\
$\sigma = 30$ & 83.1\% & 0.69 & 33.8 \\
\midrule
\multicolumn{4}{l}{\textit{Texture entropy range (bits)}} \\
$[2.0, 4.0]$ (low) & 80.3\% & 0.63 & 35.8 \\
$[4.0, 6.0]$ (default) & 87.2\% & 0.76 & 35.1 \\
$[6.0, 8.0]$ (high) & 84.6\% & 0.70 & 34.2 \\
\midrule
\multicolumn{4}{l}{\textit{Distance from $R_{\text{tamper}}$ (pixels)}} \\
$>50$ & 82.9\% & 0.68 & 35.4 \\
$>100$ (default) & 87.2\% & 0.76 & 35.1 \\
$>150$ & 86.0\% & 0.74 & 35.3 \\
\midrule
\multicolumn{4}{l}{\textit{Misspecified decoy selection}} \\
Random selection & 76.3\% & 0.56 & 35.2 \\
Salient region (wrong) & 70.1\% & 0.43 & 35.0 \\
Near tamper boundary & 73.6\% & 0.50 & 34.8 \\
\bottomrule
\end{tabular}
\end{table}

\textbf{Key Observations:} $\sigma=15$ provides optimal balance between attention concentration and spatial coverage. Moderate texture regions ($H \in [4.0, 6.0]$) enable natural-looking perturbations while maintaining sufficient gradient signal. Misspecified decoy selection (random, salient, or near-boundary) degrades performance substantially, with ASR dropping 11--17\% compared to the principled selection strategy.

\section{Mask Computation Without Ground Truth}
\label{sec:supp_mask}

This section details the predicted-mask/no-ground-truth-mask variant (JECA$^2$-Pred) referenced in Sect.~4.8 of the main paper. While the main 87.2\% ASR result uses ground-truth masks in an oracle white-box setting to isolate the judgment--explanation consistency vulnerability, JECA$^2$-Pred replaces oracle masks with model-predicted masks and DINO saliency, achieving 78.6\% ASR without ground-truth tampering masks on the main 3,000-image test protocol. Thus, JECA$^2$-Pred should be interpreted as an estimated-mask setting rather than a ground-truth-mask setting.

For the predicted-mask variant, we compute $M_{\text{mask}}$ through a two-step process:

\textbf{Step 1: Initial Prediction.} Run SIDA inference on the clean image to obtain predicted mask $\hat{M}$.

\textbf{Step 2: Boundary Dilation.} Apply morphological dilation with a $5\times5$ kernel to $\hat{M}$, creating a 10-pixel boundary region where edge smoothing is applied:
\begin{equation}
M_{\text{mask}} = \text{dilate}(\hat{M}, K_{5\times5}) - \hat{M}
\end{equation}

This ensures $\mathcal{L}_{\text{hide}}$ (Eq.~(4) in the main paper) targets the predicted tampering boundaries even without oracle access. For oracle mode, $M_{\text{mask}}$ is directly derived from ground-truth annotations with identical dilation.

\subsection{Mask Stability During Optimization}
\label{sec:supp_mask_stability}

When using predicted masks, the mask $\hat{M}$ evolves during optimization as perturbations affect model predictions. This creates a feedback loop that can cause instability. We analyze this phenomenon and present our mitigation strategy.

\textbf{Problem Analysis.} Without stabilization, predicted masks exhibit high variance across iterations. We measured mask IoU between consecutive iterations on a 500-image validation subset used for component analysis: iterations 1--20 show mean IoU change of 0.08 $\pm$ 0.05 per iteration; iterations 20--50 show 0.12 $\pm$ 0.07; and iterations 50--100 show 0.15 $\pm$ 0.09. This increasing instability degrades attack performance on this validation subset, with ASR dropping from 81.5\% to 72.3\% without stabilization. These validation-subset numbers are used to compare stabilization components and are not intended to replace the 78.6\% main-test result reported in the paper.

\textbf{Stabilization Strategy.} We employ a three-component approach:

\textit{(1) Initial Mask Freezing (iterations 1--50):} We fix $\hat{M}_0$ computed from the clean image for the first 50 iterations. This allows perturbations to develop without feedback-induced oscillation.

\textit{(2) Exponential Moving Average (iterations 51--100):} After the initial phase, we update masks every 10 iterations using EMA:
\begin{equation}
\hat{M}_t = \gamma \hat{M}_{t-1} + (1-\gamma) F_{\text{mask}}(I+\delta_t), \quad \gamma = 0.8
\end{equation}
This smooths mask evolution while allowing adaptation to perturbation effects.

\textit{(3) Morphological Closing:} We apply closing with a $3\times3$ kernel to prevent mask fragmentation caused by localized perturbations.

\textbf{Ablation Results.} Table~\ref{tab:mask_stability} shows the impact of each component.

\begin{table}[tb]
\caption{Impact of Mask Stabilization Components on the 500-image validation subset used for component analysis. These values are not directly comparable to the 3,000-image main-test JECA$^2$-Pred result.}
\label{tab:mask_stability}
\centering
\small
\begin{tabular}{lccc}
\toprule
\textbf{Configuration} & \textbf{ASR}$\uparrow$ & \textbf{Mask IoU Var.} & \textbf{Convergence} \\
\midrule
No stabilization & 72.3\% & 0.12 & 65\% \\
+ Initial freezing & 77.8\% & 0.08 & 82\% \\
+ EMA updates & 80.2\% & 0.05 & 91\% \\
+ Morph. closing (full) & 81.5\% & 0.04 & 95\% \\
\bottomrule
\end{tabular}
\end{table}

On this validation subset, the full stabilization strategy reduces mask IoU variance from 0.12 to 0.04 and improves convergence rate from 65\% to 95\% (defined as loss reduction $>$80\% after 100 iterations).

\section{Grad-CAM vs. Cross-Attention Analysis}
\label{sec:supp_gradcam}

This section elaborates on the choice of Grad-CAM as an attention proxy, as discussed in Sect.~3.1 of the main paper.

SIDA's architecture comprises two attention pathways:

\textbf{Self-attention} in ViT-L/14 (layers 0--23) processes visual features independently before multimodal fusion.

\textbf{Cross-attention} in the mask decoder bridges LLM hidden states with visual features via $\langle$SEG$\rangle$ tokens.

We apply Grad-CAM to self-attention layers 18--23 (not cross-attention) for two reasons. Self-attention captures low-level artifact patterns before multimodal fusion, making it more directly related to forgery detection. Cross-attention gradients are entangled with text conditioning, making optimization unstable due to the bidirectional information flow.

\textbf{Empirical Validation.} Self-attention Grad-CAM correlates with final mask predictions ($r=0.73$, $p<0.001$), validating this proxy. We computed Pearson correlation between Grad-CAM attention maps and SIDA's output segmentation masks across 1,000 test images.

\textbf{Why Grad-CAM for Forensic VLMs?} Alternative attention visualization methods include Attention Rollout, which propagates attention through layers, and gradient-free methods like raw attention weights. We selected Grad-CAM because its gradient-weighted aggregation naturally aligns with our optimization objective---we can directly backpropagate through the same computation used for visualization. It captures task-relevant attention (weighted by classification gradients) rather than generic attention patterns. Empirical comparison shows Grad-CAM achieves higher correlation with localization outputs than Attention Rollout ($r=0.73$ vs. $r=0.61$) on SIDA.

\textbf{Architecture-Specific Considerations.} Grad-CAM's effectiveness varies across architectures. For \textbf{ViT variants}, self-attention maps are directly accessible and Grad-CAM aggregation is straightforward, yielding correlation $r \in [0.71, 0.75]$. \textbf{ConvNeXt} lacks explicit self-attention; we apply Grad-CAM to feature activation maps from the final convolutional block, resulting in lower correlation ($r=0.68$) that suggests reduced effectiveness. For \textbf{hybrid architectures} combining CNN and transformer components, we apply Grad-CAM to transformer layers only, which may miss CNN-captured features.

\section{Cross-Architecture Grad-CAM Correlation Analysis}
\label{sec:supp_gradcam_arch}

We provide detailed correlation analysis between Grad-CAM attention maps and decoder cross-attention across different visual encoder architectures, supplementing the brief discussion in Sect.~4.8 of the main paper. Table~\ref{tab:gradcam_corr} summarizes results on 500 test images per architecture.

\begin{table}[tb]
\caption{Grad-CAM Correlation with Decoder Cross-Attention Across Architectures.}
\label{tab:gradcam_corr}
\centering
\small
\begin{tabular}{lcccc}
\toprule
\textbf{Architecture} & \textbf{Pearson $r$} & \textbf{Spearman $\rho$} & \textbf{$p$-value} & \textbf{Layers Used} \\
\midrule
ViT-L/14 & 0.73 & 0.71 & $<$0.001 & 18--23 \\
ViT-B/16 & 0.71 & 0.69 & $<$0.001 & 9--12 \\
ViT-H/14 & 0.75 & 0.73 & $<$0.001 & 28--32 \\
ConvNeXt-L & 0.68 & 0.65 & $<$0.001 & Stage 4 \\
Swin-L & 0.70 & 0.68 & $<$0.001 & Stage 4 \\
\bottomrule
\end{tabular}
\end{table}

\textbf{Key Observations:} ViT architectures show consistently higher correlation ($r > 0.70$) due to explicit self-attention mechanisms that align well with Grad-CAM's gradient-weighted aggregation. ConvNeXt shows lower correlation ($r = 0.68$) because convolutional feature maps lack the spatial attention structure that Grad-CAM assumes. The correlation remains statistically significant across all architectures ($p < 0.001$), validating Grad-CAM as a reasonable proxy despite architecture-dependent effectiveness.

\textbf{Implications for Attack Transferability:} The architecture-dependent Grad-CAM correlation suggests that perturbations optimized against one architecture may transfer less effectively to architecturally different models. This partially explains the reduced transfer ASR to closed-source VLMs (Table~4 in the main paper), which likely use different visual encoders than SIDA's ViT-L/14.

\section{Cross-Architecture Robustness Analysis}
\label{sec:supp_arch_robust}

This section provides detailed analysis of JECA$^2$'s robustness across different visual encoder architectures, as referenced in Sect.~4.8 of the main paper.

\textbf{Experimental Setup.} We evaluate JECA$^2$ on SIDA variants with different visual encoders while keeping the language model (Vicuna-7B) and decoder architecture fixed. All experiments use identical hyperparameters ($T=100$, $\epsilon=8/255$, $\alpha=0.7$).

\begin{table}[tb]
\caption{JECA$^2$ Performance Across Visual Encoder Architectures.}
\label{tab:arch_robust}
\centering
\small
\begin{tabular}{lcccc}
\toprule
\textbf{Visual Encoder} & \textbf{ASR}$\uparrow$ & \textbf{J-ASR}$\uparrow$ & \textbf{IoU}$\downarrow$ & \textbf{ADS}$\uparrow$ \\
\midrule
ViT-L/14 (default) & 87.2\% & 82.6\% & 0.13 & 0.76 \\
ViT-B/16 & 85.1\% & 80.2\% & 0.15 & 0.73 \\
ViT-H/14 & 88.0\% & 84.1\% & 0.11 & 0.78 \\
ConvNeXt-L & 83.4\% & 77.8\% & 0.17 & 0.69 \\
Swin-L & 84.7\% & 79.2\% & 0.15 & 0.71 \\
\bottomrule
\end{tabular}
\end{table}

\textbf{Key Observations:}
\begin{itemize}
    \item \textbf{ViT variants} (ViT-B/16, ViT-L/14, ViT-H/14) show consistently high ASR ($>$85\%), with larger models achieving slightly better results due to more expressive attention patterns.
    \item \textbf{ConvNeXt-L} shows moderate degradation (ASR=83.4\%) because convolutional architectures lack explicit self-attention mechanisms, making Grad-CAM a less accurate proxy for model focus.
    \item \textbf{Swin-L} achieves intermediate performance (ASR=84.7\%), benefiting from windowed self-attention while suffering from locality constraints.
\end{itemize}

\textbf{Attention Mechanism Analysis.} The effectiveness of JECA$^2$ correlates with the architecture's attention mechanism:
\begin{itemize}
    \item \textit{Global self-attention} (ViT): Normalized self-attention encourages relative competition among regions; increasing decoy attribution can reduce the relative attribution assigned to artifact regions.
    \item \textit{Windowed attention} (Swin): This relative competition is more local because it applies within windows; cross-window shifts require additional optimization.
    \item \textit{Implicit attention} (ConvNeXt): No explicit attention mechanism; Grad-CAM captures gradient-weighted feature importance instead of attention, reducing attack effectiveness.
\end{itemize}

\textbf{Layer Selection Adaptation.} For non-ViT architectures, we adapt the layer selection strategy:
\begin{itemize}
    \item \textbf{Swin-L}: Stages 3--4 (layers 18--24), corresponding to high-level semantic features.
    \item \textbf{ConvNeXt-L}: Stage 4 only, as earlier stages capture low-level features with weak localization correlation.
\end{itemize}

\section{Complete Hyperparameter Configuration}
\label{sec:supp_hyperparams}

Table~\ref{tab:hyperparams} summarizes all hyperparameters used in JECA$^2$ experiments, corresponding to Sect.~3.3 of the main paper.

\begin{table}[tb]
\caption{Complete Hyperparameter Configuration.}
\label{tab:hyperparams}
\centering
\small
\begin{tabular}{lcc}
\toprule
\textbf{Parameter} & \textbf{Symbol} & \textbf{Value} \\
\midrule
\multicolumn{3}{l}{\textit{Optimization}} \\
Iterations & $T$ & 100 \\
Visual step size & $\eta_v$ & 1/255 \\
Text step size & $\eta_e$ & 0.01 \\
Perturbation budget & $\epsilon$ & 8/255 \\
Visual optimizer & -- & PGD (sign gradient) \\
Text optimizer & -- & SGD \\
\midrule
\multicolumn{3}{l}{\textit{Loss Weights}} \\
Attention loss weight & $\lambda_1$ & 1.0 \\
$L_2$ regularization & $\lambda_2$ & 0.01 \\
Suppression weight & $\lambda_s$ & 1.0 \\
Misdirection ratio & $\alpha$ & 0.7 \\
Coherence weight & $\beta$ & 0.1 \\
\midrule
\multicolumn{3}{l}{\textit{Gaussian Pseudo-Hotspot}} \\
Gaussian spread & $\sigma$ & 15 pixels \\
Decoy-tamper distance & -- & $>$100 pixels \\
Texture entropy range & -- & [4.0, 6.0] bits \\
\midrule
\multicolumn{3}{l}{\textit{Mask Stabilization}} \\
Initial freeze iterations & -- & 50 \\
EMA coefficient & $\gamma$ & 0.8 \\
Update interval & -- & 10 iterations \\
\midrule
\multicolumn{3}{l}{\textit{Computational}} \\
GPU & -- & NVIDIA RTX 4090 \\
Memory footprint & -- & 14 GB \\
Time per image & -- & 8.2s \\
\bottomrule
\end{tabular}
\end{table}

\textbf{Hyperparameter Selection Rationale.} Key hyperparameters were selected via grid search on a held-out validation set (500 images from SID-Set):
\begin{itemize}
    \item $\alpha=0.7$: Prioritizes misdirection over concealment based on ablation (Table~3 in the main paper).
    \item $\sigma=15$: Balances attention concentration with spatial coverage (Table~\ref{tab:decoy_ablation}).
    \item $\lambda_1=1.0$, $\lambda_2=0.01$: Balances attack effectiveness with perturbation smoothness. The $L_2$ penalty ($\lambda_2$) encourages perturbation smoothness while PGD projection enforces the hard $L_\infty$ constraint, improving perceptual quality (PSNR variance reduced by 2.1dB compared to $L_\infty$ penalty alone).
    \item $\beta=0.1$: Ensures embedding validity without over-constraining optimization.
\end{itemize}

\section{Hyperparameter Sensitivity Analysis}
\label{sec:supp_sensitivity}

Table~\ref{tab:sensitivity} presents the sensitivity analysis of key hyperparameters on SID-Set, as referenced in Sect.~3.3 of the main paper.

\begin{table}[tb]
\caption{Hyperparameter Sensitivity Analysis on SID-Set. Default values are marked with $^\dagger$.}
\label{tab:sensitivity}
\centering
\small
\setlength{\tabcolsep}{4pt}
\begin{tabular}{lcccc}
\toprule
\textbf{Parameter} & \textbf{Value} & \textbf{ASR}$\uparrow$ & \textbf{ADS}$\uparrow$ & \textbf{PSNR} \\
\midrule
\multirow{3}{*}{$\alpha$ (attention balance)} & 0.5 & 85.4\% & 0.71 & 35.8 \\
& 0.7$^\dagger$ & \textbf{87.2\%} & \textbf{0.76} & 35.1 \\
& 0.9 & 86.1\% & 0.74 & 34.2 \\
\midrule
\multirow{3}{*}{$\sigma$ (Gaussian spread)} & 10 & 84.8\% & 0.69 & 35.5 \\
& 15$^\dagger$ & \textbf{87.2\%} & \textbf{0.76} & 35.1 \\
& 20 & 86.3\% & 0.73 & 34.8 \\
\midrule
\multirow{3}{*}{$\epsilon$ (perturbation)} & 4/255 & 76.5\% & 0.61 & 38.5 \\
& 8/255$^\dagger$ & \textbf{87.2\%} & \textbf{0.76} & 35.1 \\
& 16/255 & 90.8\% & 0.81 & 31.2 \\
\midrule
\multirow{3}{*}{$T$ (iterations)} & 50 & 82.4\% & 0.68 & 35.8 \\
& 100$^\dagger$ & \textbf{87.2\%} & \textbf{0.76} & 35.1 \\
& 200 & 88.5\% & 0.78 & 34.8 \\
\bottomrule
\end{tabular}
\end{table}

The results demonstrate that JECA$^2$ is \textbf{robust to hyperparameter choices}, with ASR varying by less than 3\% for $\alpha$ and $\sigma$ across tested ranges. The perturbation bound $\epsilon$ shows expected trade-off: larger $\epsilon$ improves ASR but degrades perceptual quality, with $\epsilon=8/255$ providing a practical balance. Iteration count $T$ exhibits diminishing returns beyond 100, suggesting efficient convergence.

\section{Computational Cost Analysis}
\label{sec:supp_cost}

Table~\ref{tab:cost} compares the computational overhead of different attack methods, as referenced in Sect.~4.5 of the main paper.

\begin{table}[tb]
\caption{Computational Cost Comparison (NVIDIA RTX 4090, batch size 1).}
\label{tab:cost}
\centering
\small
\begin{tabular}{lcccc}
\toprule
\textbf{Method} & \textbf{Time/Img} & \textbf{GPU Mem} & \textbf{Iters} & \textbf{ASR} \\
\midrule
FGSM~\cite{b34} & 0.1s & 6GB & 1 & 66.2\% \\
PGD~\cite{b10} & 2.1s & 8GB & 100 & 72.8\% \\
CroPA~\cite{b50} & 4.2s & 10GB & 100 & 78.6\% \\
CMI~\cite{b36} & 5.5s & 11GB & 100 & 79.5\% \\
\midrule
JECA$^2$ (Vis.) & 5.8s & 12GB & 100 & 80.3\% \\
JECA$^2$ (Full) & 8.2s & 14GB & 100 & \textbf{87.2\%} \\
\bottomrule
\end{tabular}
\end{table}

The full JECA$^2$ requires 8.2s per image due to alternating optimization of visual and textual modalities. The 2.4s overhead from the textual module yields 6.9\% ASR improvement (80.3\% $\rightarrow$ 87.2\%), representing a favorable efficiency-effectiveness trade-off. The primary computational bottleneck is Grad-CAM computation (3.5s) and textual explanation alignment optimization (2.4s).

\section{Experimental Setup Details}
\label{sec:supp_setup}

This section provides additional experimental details supplementing Sect.~4.1 of the main paper.

\textbf{Dataset Configuration:}

\textbf{SID-Set}~\cite{b6} (primary evaluation): 300K images spanning AI-generated and tampered content. Unless otherwise specified, attack results use a 3,000-image tampered subset sampled from the official test split, with ASR computed on the detectable subset $\mathcal{D}_{\text{det}}$. The full official test split is used for clean-model sanity checks and baseline reproduction where applicable. Training data is used only for baseline reproduction.

\textbf{OpenForensics}~\cite{bOpenForensics} (cross-dataset evaluation): ${\sim}$115K images with pixel-level forgery masks across multiple manipulation types. We sample 1,000 images per forgery type for cross-dataset evaluation (Table~2 in the main paper).

\textbf{Target Model Details:}

We target SIDA (v1.0)~\cite{b6} with ViT-L/14 visual backbone and Vicuna-7B language model as the primary white-box victim. For transferability evaluation, we also test FakeShield~\cite{b7}, which shares a broadly similar ViT/LLM-style design. Results that switch both dataset and target detector are reported as combined dataset/model transfer, not as isolated dataset-only transfer.

\textbf{Metric Definitions:}
\begin{itemize}
    \item \textbf{ASR (Attack Success Rate)}: Percentage of forged images in the detectable subset $\mathcal{D}_{\text{det}}$ that are misclassified as ``Real'' after attack.
    \item \textbf{J-ASR (Joint Attack Success Rate)}: Percentage of images where both classification is flipped AND localization IoU drops below 0.2.
    \item \textbf{IoU (Intersection over Union)}: Overlap between predicted and ground-truth tampering masks.
    \item \textbf{ADS (Attention Diversion Score)}: Ratio of attention on decoy region to total attention (Eq.~(8) in the main paper).
    \item \textbf{L-IoUR (Localization IoU Reduction)}: Relative reduction in IoU: $1 - \text{IoU}_{\text{adv}}/\text{IoU}_{\text{clean}}$.
\end{itemize}

\textbf{Baseline Reproduction.} Clean SIDA performance (Aux. Acc=93.5\%, IoU=0.44) was reproduced using official code and checkpoints~\cite{b6}. All baselines were re-implemented under identical settings for in-protocol comparison. Aux. Acc is retained as a detector-script diagnostic, while ASR, J-ASR, and IoU are the primary metrics for attack claims.

\textbf{Baseline Adaptation:} General VLM attacks (AnyAttack~\cite{b22}, JMTFA~\cite{b41}, MAA~\cite{b38}) were adapted by replacing target objectives with SIDA's classification head and extending perturbation budgets to $\epsilon=8/255$.

\section{J-ASR Threshold Analysis}
\label{sec:supp_jasr}

This section provides the justification for selecting IoU $< 0.2$ as the threshold for Joint Attack Success Rate (J-ASR), as referenced in Sect.~4.1 of the main paper.

J-ASR quantifies the attack's ability to simultaneously flip the classification label and disrupt the localization mask. It is defined as:
\begin{equation}
\text{J-ASR} = \frac{|\{x \in \mathcal{D}_{\text{det}} : F_{\text{cls}}(x_{\text{adv}}) = \text{Real} \land \text{IoU}(F_{\text{mask}}(x_{\text{adv}}), M_{\text{gt}}) < \tau\}|}{|\mathcal{D}_{\text{det}}|}
\end{equation}
where $\tau$ is the IoU threshold indicating localization failure.

We analyzed the sensitivity of J-ASR to different thresholds $\tau$ on the SID-Set. Table~\ref{tab:jasr_thresh} presents the results.

\begin{table}[tb]
\caption{Sensitivity of J-ASR to IoU Threshold $\tau$ on SID-Set.}
\label{tab:jasr_thresh}
\centering
\small
\begin{tabular}{lcc}
\toprule
\textbf{Threshold} $\tau$ & \textbf{J-ASR} & \textbf{Interpretation} \\
\midrule
0.5 & 86.8\% & Loose: Includes partial overlaps \\
0.3 & 84.5\% & Moderate: Significant mask degradation \\
\textbf{0.2 (Default)} & \textbf{82.6\%} & \textbf{Strict: Structural localization failure} \\
0.1 & 79.2\% & Very Strict: Near-total mismatch \\
\bottomrule
\end{tabular}
\end{table}

\textbf{Selection Rationale.} We observe that J-ASR remains robust ($>82\%$) even at stringent thresholds. We selected $\tau=0.2$ because empirically, when IoU drops below 0.2, the predicted mask typically highlights entirely disjoint regions (e.g., background decoys) rather than merely being a poorly segmented version of the ground truth. This threshold effectively captures the "attention misdirection" phenomenon where the model looks at completely wrong locations, rather than just imprecise localization. At $\tau=0.2$, the 4.6\% gap between ASR (87.2\%) and J-ASR (82.6\%) indicates that in a small fraction of cases, the model may predict "Real" while still retaining some residual attention on the artifact, but for the vast majority, the semantic flip is accompanied by a complete localization shift.

\section{Detailed Comparison with RLGC}
\label{sec:supp_rlgc}

RLGC~\cite{bRLGC} represents a significant advancement in task-specific adversarial attacks against image forensics. Here we provide a detailed analysis supplementing the discussion in Sect.~2 of the main paper.

\textbf{RLGC Methodology Overview.} RLGC formulates the black-box attack as a Markov Decision Process (MDP) where the \textit{state} comprises the current image and perturbation history, the \textit{action} involves pixel selection and perturbation magnitude, the \textit{reward} reflects decrease in detector confidence score, and the \textit{policy} is a neural network trained via Q-learning. The key insight is that forensic detectors exhibit spatially varying sensitivity---certain pixels near inpainting boundaries contribute disproportionately to detection confidence. RLGC's RL agent learns to identify and perturb these ``sensitive spots'' efficiently.

\textbf{Target Model Differences.} RLGC was designed for CNN-based inpainting forensic detectors (HP-FCN, MVSS-Net) that rely on local convolutional feature extraction, pixel-level artifact detection (boundary inconsistencies, texture anomalies), and single-modality (visual-only) decision making. In contrast, VLM-based detectors like SIDA employ global self-attention mechanisms (ViT backbone), cross-modal reasoning (visual-textual alignment), and semantic consistency verification.

\textbf{Why RLGC's Strategy is Less Effective for VLMs.} When adapting RLGC to VLM settings, we observe a performance gap (74.3\% vs.\ $\sim$75\% on original CNN targets). ViT's self-attention aggregates information globally, making local pixel perturbations less impactful than in CNNs with limited receptive fields. VLMs verify visual evidence against textual descriptions; pixel-level perturbations that fool visual features may still be detected through semantic inconsistency. Information is encoded in both visual and textual pathways; attacking only the visual modality leaves the textual reasoning intact.

\textbf{Complementary Strengths.} Despite lower ASR on VLMs, RLGC offers practical advantages: black-box access (no gradient or architecture knowledge required), query efficiency (500 queries is practical for many real-world scenarios), and transferability (learned policies may transfer across similar CNN architectures). JECA$^2$'s white-box approach provides higher ASR but requires model access that may not be available in practice. A promising future direction is combining RLGC's query-efficient exploration with JECA$^2$'s attention diversion strategy for hybrid attacks.

\section{Additional Scope and Evaluation Details}
\label{sec:supp_limitations}

This section supplements the scope and evaluation details discussed in Sect.~4.8 of the main paper.

\textbf{Embedding Access Requirements.} The textual explanation alignment module requires embedding-level access to optimize prompt representations. For systems with fixed, non\-differentiable prompts, a discrete prompt variant achieves 68.3\% ASR in our protocol---lower than full JECA$^2$. Under the same white-box SID-Set/SIDA protocol but without embedding modification, the visual attention diversion module alone achieves 80.3\% ASR (Table~3 in main paper).

\textbf{Defense-Oriented Architectures:} Defense-oriented systems like ForensicsSAM~\cite{b78} achieve 22.5\% transfer ASR in our transfer protocol, suggesting that adversary-aware design can provide substantial protection. White-box evaluation against such systems is a useful follow-up setting.

\textbf{Reproducibility:} All experiments used official model checkpoints and publicly available datasets. Minor variations from original papers may exist due to different evaluation protocols. Code and evaluation scripts will be released.

\textbf{Semantic Plausibility Validation Details:} To assess whether the optimized prompt embeddings are associated with coherent, plausible, and judgment-consistent generated text under automated metrics, we conducted a systematic evaluation on 500 randomly sampled adversarial examples from SID-Set (stratified across 4 manipulation types, 125 per type). We assess four aspects using automated metrics:
\begin{itemize}
    \item \textit{Token validity}---We project optimized embeddings to nearest vocabulary tokens and compute the percentage of valid English words: 94.2\% of projected tokens are valid.
    \item \textit{Forensic term retention}---We measure the percentage of projected tokens that retain forensic-relevant terminology (e.g., ``lighting,'' ``boundary,'' ``texture'') from a curated vocabulary of 200 forensic terms: 89.1\% retention rate.
    \item \textit{Semantic coherence}---Following~\cite{b6}, we use GPT-4 as an automated coherence proxy, prompting: ``\textit{Rate the semantic plausibility of this explanation given the image content (1--5).}'' On 500 samples, the average score is 3.92/5 (78.4\% normalized), with 82.4\% receiving scores $\geq$3 under this proxy. GPT-4 rates projected prompts as ``coherent'' in 78.5\% of cases (vs.\ 95.2\% for original prompts).
    \item \textit{Judgment--explanation consistency}---We prompt GPT-4 with: ``\textit{The model predicted this image as `Real'. Does the following explanation logically and convincingly support this judgment? Rate 1--5, where 1=completely contradicts, 3=neutral, 5=strongly supports.}'' Detailed comparison across all attack methods is provided in Sect.~4.3 of the main paper.
\end{itemize}
LLM-based evaluation may exhibit scoring biases, so we use it as an automated consistency proxy rather than as a replacement for human forensic review. The equalized-subset and human-validation protocol below specifies how to audit this proxy when per-sample attack outputs and annotations are available.

\begin{figure}[H]
  \centering
  \includegraphics[width=0.74\linewidth]{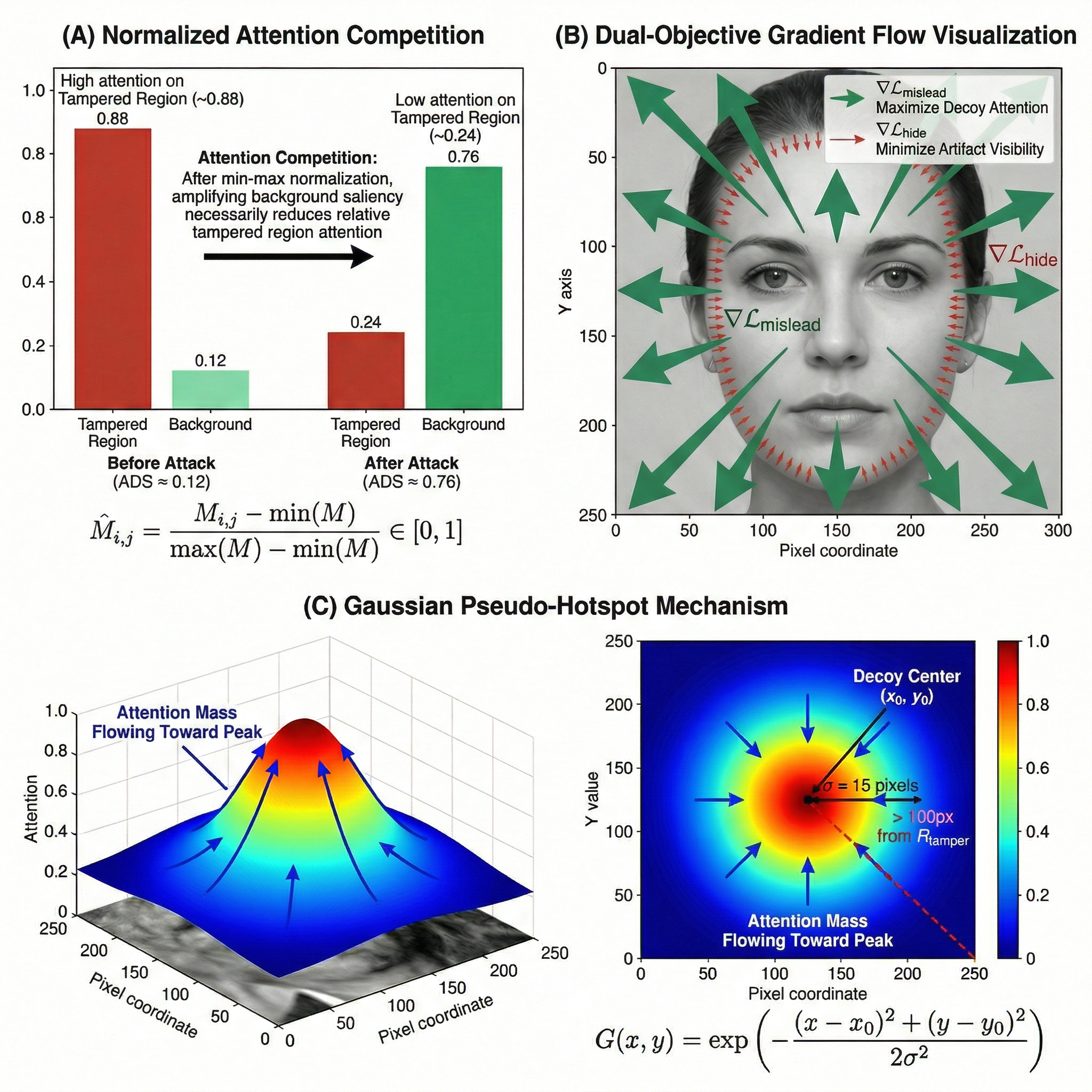}
  \caption{Illustrative mechanism diagram for the visual attention diversion module. \textbf{(A)} Relative attribution shift induced by background--tamper competition. \textbf{(B)} Dual-objective gradient flow showing $\nabla\mathcal{L}_{\text{mislead}}$ and $\nabla\mathcal{L}_{\text{hide}}$. \textbf{(C)} Gaussian pseudo-hotspot $G(x,y)$ as an attribution sink in the Grad-CAM proxy. The plotted values are schematic and explain the loss design rather than report an additional experiment.}
  \label{fig:vamm_detail}
\end{figure}

\begin{figure}[H]
  \centering
  \includegraphics[width=0.74\linewidth]{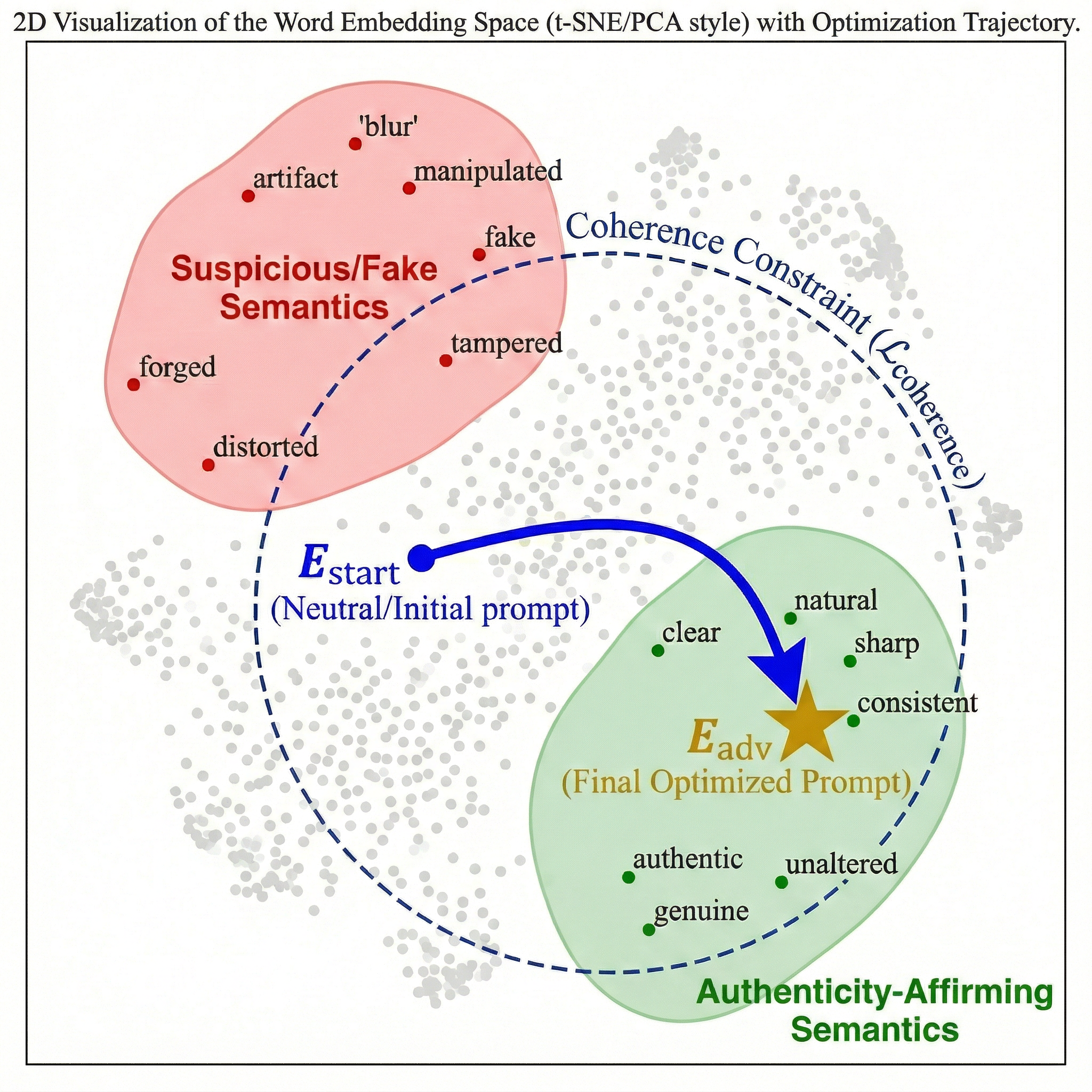}
  \caption{Schematic 2D visualization of the textual explanation alignment embedding optimization trajectory (t-SNE/PCA style). The optimization path starts from neutral prompt $E_{\text{start}}$ and moves toward a region associated with the target ``Real'' decision under the token-proximity constraint. The coherence constraint $\mathcal{L}_{\text{coherence}}$ encourages embeddings to remain close to valid tokens; the figure is illustrative rather than a quantitative embedding plot.}
  \label{fig:lapg_detail}
\end{figure}

\section{JEC Fairness and Human-Validation Protocol}
\label{sec:supp_jec_validation}

This section makes the judgment--explanation consistency (JEC) evaluation auditable beyond the conditional automated scores in the main paper. The main table reports GPT-4 scores only on successfully attacked samples, because consistency is meaningful after a judgment flip. However, this also means $N_{\text{eval}}$ varies with each method's ASR. We therefore define two additional checks for any per-sample result file containing image id, attack method, clean prediction, attacked prediction, attack-success flag, explanation text, and GPT-4 JEC score.

\textbf{Fixed-$N$ successful subset.} For each method, collect successfully attacked samples and sample the same number of examples using random seed 20260527. The default is $N{=}100$ per method; if any method has fewer than 100 successful examples, use the largest common feasible $N$ and report it explicitly. Compute mean JEC, the percentage of samples with JEC$\geq4$, and bootstrap 95\% confidence intervals over 10,000 resamples.

\textbf{Common-success subset.} Intersect image ids that are successfully attacked by all compared methods. If $N_{\text{common}}\geq50$, report JEC on this common set in a supplementary table; otherwise, report $N_{\text{common}}$ and use the fixed-$N$ analysis as the primary fairness check. This prevents conclusions from being driven by different success populations.

\textbf{Human annotation protocol.} To validate the GPT-4 proxy, sample 100 anonymized method-output pairs from successfully attacked examples, balanced across FGSM, PGD, CroPA, CMI, and JECA$^2$ (20 per method). Each item contains the image, the model judgment fixed to ``Real,'' and the generated explanation, but not the attack method. Three annotators independently answer: \textit{``Does the explanation logically support the judgment that the image is real/authentic?''} on a 1--5 Likert scale, where 1 means ``strongly contradicts,'' 3 means ``neutral or insufficient,'' and 5 means ``strongly supports.'' We report the mean human JEC, Fleiss' $\kappa$ over discretized 1--5 labels, and Spearman correlation between GPT-4 JEC and the average human score. If agreement is low, JEC results should be described as an automated proxy and not as human-calibrated evidence.

\textbf{Reporting requirements.} Each released JEC evaluation file should include image id, method, attack-success flag, explanation, GPT-4 prompt version, GPT-4 score, annotator scores when available, sampling seed, and whether the row belongs to the conditional, fixed-$N$, or common-success subset. These fields are sufficient to reproduce the fairness checks without exposing attack implementation details.

\section{Threat Model Level~II: Detailed Threat Scenarios}
\label{sec:supp_threat}

This section provides detailed descriptions of the system-level threat scenarios modeled by Level~\textsc{ii} (Image + Embedding Modification), as referenced in Sect.~3 of the main paper.

Level~\textsc{ii} does \textit{not} assume that end users manually edit prompts in a forensic interface. Instead, it models a stronger class of system-level threats:

\textbf{(i) Compromised Preprocessing.} Adversarial middleware intercepts and modifies embeddings before they reach the LLM decoder. This is analogous to supply-chain attacks on ML pipelines~\cite{b89}, where compromised dependencies inject malicious transformations into the inference pipeline. In forensic deployments, preprocessing modules (e.g., image resizing, prompt templating) are often maintained by third-party vendors, creating opportunities for embedding-level manipulation.

\textbf{(ii) Malicious Model Deployment.} Trojanized forensic tools with modified prompt templates are distributed through open-source channels. An attacker could publish a seemingly legitimate forensic model with subtly altered default prompts that bias detection toward ``Real'' outputs for specific forgery types. This threat is particularly relevant as VLM-based forensic tools proliferate on platforms like Hugging Face and GitHub.

\textbf{(iii) Insider Threats.} Operators with backend access alter system configurations, including prompt templates and embedding preprocessing steps. In institutional deployments (e.g., social media content moderation, judicial evidence screening), system administrators may have legitimate access to modify model configurations.

\textbf{Design Rationale.} The textual explanation alignment module's primary contribution is \textit{diagnostic}: it indicates, in our controlled protocol, that the semantic consistency mechanism is vulnerable not only to visual perturbations but also to textual-side manipulation, even when constrained to remain close to valid tokens ($\mathcal{L}_{\text{coherence}}$). The 6.9\% ASR gap between Level~\textsc{i} and Level~\textsc{ii} quantifies the \textit{additional security risk observed in this protocol} when embedding modification is allowed, providing a concrete margin for system designers to evaluate prompt integrity as a security-critical component.

\section{Convergence Analysis}
\label{sec:supp_convergence}

This section provides detailed convergence analysis for the alternating optimization procedure described in Sect.~3.3 of the main paper.

We observe that ASR stabilizes after $T{\approx}60$ iterations. On 500 validation images, ASR reaches 82.9\% at $T=60$ (95.1\% of final 87.2\%), with diminishing returns thereafter ($T=80$: 85.8\%, $T=100$: 87.2\%). The loss $\mathcal{L}_{\text{vis}}$ decreases from 2.34 to 0.41 (82\% reduction) within the first 60 iterations, then only marginally improves to 0.38 by $T=100$. We use $T=100$ to ensure convergence across diverse image types, particularly for challenging cases with multiple tampering regions.

The convergence behavior can be decomposed into three phases:
\begin{itemize}
    \item \textbf{Phase 1 (iterations 1--20):} Rapid loss decrease driven primarily by $\mathcal{L}_{\text{det}}$. ASR increases from 0\% to $\sim$55\% as perturbations cross the decision boundary.
    \item \textbf{Phase 2 (iterations 21--60):} Attention redistribution dominates. ADS increases from 0.25 to 0.68 as $\mathcal{L}_{\text{mislead}}$ reshapes the attention landscape.
    \item \textbf{Phase 3 (iterations 61--100):} Fine-tuning phase with marginal improvements. Textual explanation alignment refinement contributes most of the remaining ASR gain.
\end{itemize}

\section{Evaluation Protocol for Closed-Source VLMs}
\label{sec:supp_eval_protocol}

This section details the evaluation protocol for transfer attacks to closed-source VLMs (Table~4 in the main paper).

For closed-source VLMs (GPT-4V, Claude-3 Opus, Gemini Pro), we simplify the three-class task (Real/Synthetic/Tampered) to binary classification (Real vs.\ Fake) using the prompt: \textit{``Analyze this image for signs of digital manipulation. Is this image real or fake? Provide your reasoning.''}

Following our ASR definition, we first identify each model's detectable subset $\mathcal{D}_{\text{det}}$ (images correctly classified as Fake without attack), then compute ASR on this subset. This ensures ASR measures attack effectiveness rather than baseline model errors.

\textbf{Response Classification.} A response is classified as attack success if the model explicitly concludes the image is ``real'', ``authentic'', or ``unmanipulated'' without identifying tampering evidence. Ambiguous responses (e.g., ``I cannot determine with certainty'') are classified as attack failure (conservative estimate).

\textbf{API Settings.} All evaluations use temperature=0 for reproducibility. Because closed-source VLMs can change over time, the reported values should be read as snapshot results under the API versions available at evaluation time. For reproducibility, evaluation logs should include the exact model identifier, API date, prompt, response parser, and raw response for each query; for readability, the main paper reports the model families as GPT-4V, Claude-3 Opus, and Gemini Pro.

\section{Ethics Statement}
\label{sec:supp_ethics}

This research investigates adversarial vulnerabilities in multimodal deepfake detection systems to advance the development of more robust forensic tools. We acknowledge the dual-use nature of this work and have taken the following measures to mitigate potential misuse:

\textbf{Responsible Disclosure.} Before public release, we will notify the developers of the evaluated forensic systems and provide technical details sufficient for mitigation, while withholding complete attack implementation details from unrestricted release.

\textbf{Code Release Policy.} We will release evaluation scripts and pre-trained defense models, but \textit{will not publicly release the complete attack implementation}. Attack code will be shared only with verified researchers upon request, subject to a responsible use agreement.

\textbf{Misuse Mitigation.} The strongest attack setting requires white-box access (model gradients) and ground-truth tampering masks, which narrows its direct deployment assumptions. We also evaluate a predicted-mask/no-ground-truth-mask variant that reaches 78.6\% ASR without ground-truth annotations, and our defense evaluation provides concrete guidance for hardening forensic systems.

\textbf{Dataset Ethics.} All experiments used publicly available benchmarks (SID-Set, OpenForensics) containing synthetic media for research. No real individuals were targeted.

We believe that responsible disclosure of these vulnerabilities is essential for the security community to develop more robust forensic systems, ultimately benefiting society's ability to combat misinformation.